\pdfoutput=1
\documentclass[11pt]{article}

\usepackage[]{acl}

\usepackage{times}
\usepackage{latexsym}
\usepackage[T1]{fontenc}
\usepackage[utf8]{inputenc}

\usepackage{microtype}

\usepackage{inconsolata}
\usepackage[T1]{fontenc}    % use 8-bit T1 fonts
\usepackage{hyperref}       % hyperlinks
\usepackage{url}            % simple URL typesetting
\usepackage{booktabs}       % professional-quality tables
\usepackage{amsfonts}       % blackboard math symbols
\usepackage{nicefrac}       % compact symbols for 1/2, etc.
\usepackage{microtype}      % microtypography
\usepackage{graphicx}
\usepackage{wrapfig, blindtext, booktabs}
\usepackage{tabulary}
\usepackage{caption}
\usepackage{subcaption}
\usepackage[most]{tcolorbox}
\tcbuselibrary{listings}
\newtcblisting{codeblock}[1]{title=#1}
\usepackage{longtable}
\usepackage{amsmath}
\usepackage{enumitem}
\usepackage{breqn}
\usepackage[ruled]{algorithm2e}
\usepackage{algorithmicx}
\usepackage{multirow}
\newcommand\blfootnote[1]{%
\begingroup
\renewcommand\thefootnote{}\footnote{#1}%
\addtocounter{footnote}{-1}%
\endgroup
}
\usepackage{comment}

\newcount\Comments  % 0 suppresses notes to selves in text
\definecolor{purple}{rgb}{0.5,0,1}
\definecolor{teal}{rgb}{0.33,0.65,0.55}
\definecolor{green}{rgb}{0.1,0.65,0.1}
\newcommand{\kibitz}[2]{\ifnum\Comments=1\textcolor{#1}{#2}\fi}

\title{SYNFAC-EDIT: Synthetic Imitation Edit Feedback for \\ Factual Alignment in Clinical Summarization}

\author{%
  Prakamya Mishra~\thanks{indicates equal contribution}\thanks{Presently in AMD AI}$^{1,3}$, 
  Zonghai Yao~\footnotemark[1]$^{1}$ \\
  \bf{Parth Vashisht}$^{1}$,  
  \bf{Feiyun Ouyang}$^{3}$, 
  \bf{Beining Wang}$^{2}$, 
  \bf{Vidhi Dhaval Mody}$^{1}$, 
  \bf{Hong Yu}$^{1,3}$
  \\
  University of Massachusetts, Amherst$^1$, Fudan University$^2$\\
  University of Massachusetts, Lowell$^3$
  \\
  \texttt\ \{\href{mailto:prakamyamish@umass.edu}{prakamyamish}, \href{mailto:zonghaiyao@umass.edu}{zonghaiyao}\}@{umass.edu}
  \\
}

\begin{document}
\maketitle
\begin{abstract}
Large Language Models (LLMs) such as \texttt{GPT} \& \texttt{Llama} have demonstrated significant achievements in summarization tasks but struggle with factual inaccuracies, a critical issue in clinical NLP applications where errors could lead to serious consequences.
%misdiagnoses. 
To counter the high costs and limited availability of expert-annotated data for factual alignment, this study introduces an innovative pipeline that utilizes \texttt{>100B} parameter \texttt{GPT} variants like \texttt{GPT-3.5} \& \texttt{GPT-4} to act as synthetic experts to generate high-quality synthetics feedback aimed at enhancing factual consistency in clinical note summarization.
Our research primarily focuses on edit feedback generated by these synthetic feedback experts without additional human annotations, mirroring and optimizing the practical scenario in which medical professionals refine AI system outputs. 
Although such \texttt{100B+} parameter \texttt{GPT} variants have proven to demonstrate expertise in various clinical NLP tasks, such as the Medical Licensing Examination, there is scant research on their capacity to act as synthetic feedback experts and deliver expert-level edit feedback for improving the generation quality of weaker (\texttt{<10B} parameter) LLMs like \texttt{GPT-2 (1.5B)} \& \texttt{Llama 2 (7B)} in clinical domain.
So in this work, we leverage \texttt{100B+ GPT} variants to act as synthetic feedback experts offering expert-level edit feedback, that is used to 
reduce hallucinations and align weaker (\texttt{<10B} parameter) LLMs with medical facts using two distinct alignment algorithms (DPO \& SALT), endeavoring to narrow the divide between AI-generated content and factual accuracy.
This highlights the substantial potential of LLM-based synthetic edits in enhancing the alignment of clinical factuality~\footnote{Dataset is released here: \url{https://huggingface.co/datasets/bio-nlp-umass/SYNFAC-EDIT}.}.

\end{abstract}

\section{Introduction}

\blfootnote{$\dagger$ To appear in proceedings of the Main Conference on Empirical Methods in Natural Language Processing (EMNLP) 2024}

% How LLMs are good but have problems with Hallucination.
% Relate it to the problem being solved in the general domain and how these problems cannot be solved in clinical.

\begin{figure*}[ht]
  \centering
  % \vspace{-6mm}
    \includegraphics[width=1\textwidth]{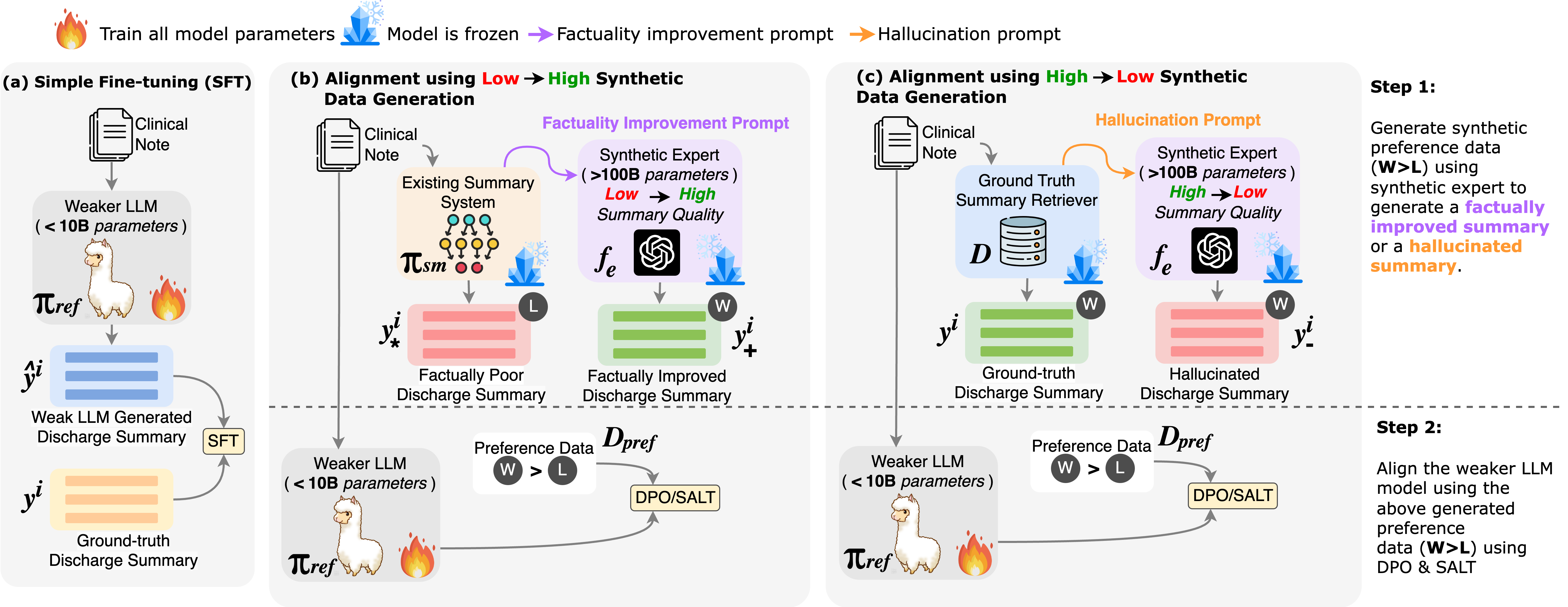}
    % \vspace{-6mm}
  \caption{\textbf{(a):} The illustration of a standard simple fine-tuning pipeline. \textbf{(b \& c):} The illustration of our proposed synthetic edit feedback generation \& alignment training pipeline. In \textbf{Step 1} of our synthetic edit feedback generation pipeline we generate preference data in two directions: (1) \texttt{\textbf{Low$\rightarrow$High}} \textbf{[b]}, where we generate a factually improved summary given an unaligned existing summary system generated summary (Section \ref{sec:L2HData}); (2) \texttt{\textbf{High$\rightarrow$Low}} \textbf{[c]}, where we generate a hallucinated summary given a clinical not article and a ground truth reference summary (Section \ref{sec:H2LData}). In \textbf{Step 2} we align the Weaker LLM model using the Step 1 generated preference data using two alignment algorithms namely DPO \& SALT (Section \ref{sec:alignment_training}).}
  \label{fig:main}
  % \vspace{-6mm}
\end{figure*}

The advent of generative artificial intelligence (AI) has been markedly accelerated by the development of large language models (LLMs) such as \texttt{GPT-3} \cite{NEURIPS2020_1457c0d6}, \texttt{GPT-4} \cite{openai2023gpt4}, \texttt{Llama} \cite{touvron2023llama}. These models have demonstrated superior capabilities in natural language understanding and natural language generation, outperforming language model predecessors (LMs) like \texttt{T5} \cite{10.5555/3455716.3455856} and \texttt{GPT-2} \cite{radford2019language} in a variety of linguistic tasks. 
Despite these advancements, LLMs confront significant challenges, primarily their propensity for generating hallucinations—fabricated information not grounded in source text—and producing factually inconsistent outputs~\cite{10.1145/3571730, zhang2023language, maynez-etal-2020-faithfulness}. Such limitations critically undermine the models' reliability, particularly critical in clinical NLP applications, where inaccuracies could result in serious misdiagnoses.

The NLP community has discussed many reasons for the hallucination problem, including some limitations stemming from traditional supervised fine-tuning (SFT). SFT fails to differentiate between significant errors, such as hallucinations, and minor inaccuracies, like grammatical mistakes, treating all errors equally in their loss calculations. Moreover, SFT applies a uniform loss weighting across all data, regardless of its type, quality, or complexity, potentially diluting the training signal for more critical learning objectives.
In response to these limitations, recent research has explored learning paradigms incorporating human feedback, such as RLHF~\cite{NEURIPS2022_b1efde53, ziegler2020finetuning, NEURIPS2020_1f89885d}, RLAIF~\cite{lee2023rlaif}, RRHF~\cite{yuan2023rrhf}, and RAFT~\cite{teed2020raft}. Techniques including PPO~\cite{baselines}, DPO~\cite{rafailov2023direct}, and SALT~\cite{yao2023improving} have demonstrated effectiveness in aligning these hallucination-prone models. 
However, these alignment methods require substantial amounts of human-annotated data to illustrate human preferences during training, which can be challenging to obtain in clinical domains \cite{li2023feasibility, yoo-etal-2021-gpt3mix-leveraging, dai2023auggpt}.

In this work, we mainly focus on previously less-studied edit feedback data to better align LMs (\texttt{GPT-2 (1.5B)}) \& LLMs (\texttt{Llama-2 (7B)}) to generate factually correct clinical note summaries. Recent works \cite{casper2023open, ji2023ai, yao2023improving} discussed some limitations of current common feedback types (comparison or rating feedback) and the advantages of adding edit feedback for better human alignment.
Human edits are a more natural way to collect feedback from clinicians as they fix AI-generated text for their workflow to improve generation~\cite{yao2023improving}. Collecting other forms of feedback that are not directly tied to the clinician’s workflow will not scale as much, this is especially true in domains requiring expert domain knowledge and with nuanced user goals. Considering the cost, time, and availability of the experts, it is important to collect edit feedback from the expert’s daily workflow.
However, it is challenging to collect real-world clinician's edit feedback due to privacy protection and strict data regulations like HIPAA ~\cite{annas2003hipaa}.

Generating a synthetic imitation edit feedback dataset by leveraging large (\texttt{>100B} parameters) \texttt{GPT} variants like \texttt{GPT-3.5} \& \texttt{GPT-4}~\cite{eysenbach2023role, li2023feasibility, Dai2023ChatAugLC} is one potential solution~\footnote{We used Azure OpenAI Service, which is HIPAA-regulated: https://azure.microsoft.com/en-us/products/ai-services/openai-service/}. Such dataset can then be used for alignment training.
Although these large \texttt{GPT} variants have reached expert-level performance in many clinical NLP tasks (e.g., Medical Licensing Examination) \cite{kung2023performance, gilson2023does,Yang2023.10.26.23297629}, there is not much previous work discussing whether they can generate expert-level edit feedback for LMs and LLMs in the clinical NLP tasks. 
Therefore, we propose to use these large GPT variants as synthetic experts for generating high-quality edit feedback for fine-tuning LLMs and LMs using the recent SOTA alignment methods like DPO \cite{rafailov2023direct} \& SALT \cite{yao2023improving} for improving factuality in the clinical domain for the clinical note summarization task.
Specifically, we propose a new pipeline to generate synthetic preference-based data in two directions for alignment training: 1) \noindent\textbf{\texttt{High$\rightarrow$Low}:} where we use these large GPT variants as synthetic experts to add factual hallucinations to generate factuality-based low-quality dispreferred summaries given the original factuality-based high-quality preferred ground-truth summaries \& the corresponding clinical notes. 2) \noindent\textbf{\texttt{Low$\rightarrow$High}:} where we use these large GPT variants as synthetic experts to add factual information to generate factuality-based high-quality preferred summaries given the factuality-based low-quality dispreferred unaligned model generated summaries and the corresponding clinical notes. We then treat the high-quality summaries as the preferred ones and the low-quality summaries as the dispreferred ones used in our synthetic preference data pairs.

Our experiments demonstrate the efficacy of utilizing synthetic edit feedback to enhance the factual accuracy of model-generated summaries. 
Specifically, for \texttt{Llama2 (7B)}, we observed a 2.44\% $\uparrow$ in \texttt{ROUGEL} and a 1.35\% $\uparrow$ in factuality using the DPO.
Similarly, SALT resulted in a 2.47\% $\uparrow$ in \texttt{ROUGEL} and a 2.04\% $\uparrow$ in factuality. 
For \texttt{GPT-2}, DPO led to a 3.04\% $\uparrow$ in \texttt{ROUGEL} and a 2.93\% $\uparrow$ in factuality, while SALT yielded a 4.04\% $\uparrow$ in \texttt{ROUGEL} and a 4.64\% $\uparrow$ in factuality. 
Moreover, our top-performing model garnered a 78\% preference rate for factuality among human evaluators, highlighting its superior performance.
% ~\footnote{Our code and dataset will be released at \url{https://github.com/seasonyao/LearnFromHumanEdit}}.

% Our experiments demonstrate the effectiveness of the synthetic edit feedback for improving factuality in the model-generated summaries: 
% for \texttt{Llama2}, a 2.44\%$\uparrow$ \texttt{ROUGEL} and 1.35\%$\uparrow$ Factuality with DPO, and a 2.47\%$\uparrow$ \texttt{ROUGEL} and 2.04\%$\uparrow$ Factuality with SALT; 
% and for \texttt{GPT2}, a 3.04\%$\uparrow$ \texttt{ROUGEL} and 2.93\%$\uparrow$ Factuality with DPO, and a 4.04\%$\uparrow$ \texttt{ROUGEL} and 4.64\%$\uparrow$ Factuality with SALT,
% compared to the SFT baseline. 
% Our best model also received a higher preference (with a 78\% win rate) for factuality by human annotators.

\section{Problem Statement}
\label{sec:problem-statement}
Given an available dataset $D\mathbin{:}\{X,Y\}$ of $C$ clinical notes $X\mathbin{:}\{x^1,x^2,...x^C\}$, their corresponding ground truth reference discharge summaries $Y:\{y^1,y^2,...y^C\}$, and a reference model $\pi_{ref}$, the aim of the clinical note summarization task $T$ is to train the model $\pi_{ref}(y^i|x^i)$. Here the $i^{th}$ clinical note $x^i\mathbin{:}\{x_1^{i},x_2^{i},...x_n^{i}\}$ consists of $n$ tokens ($j^{th}$ token represented by $x_j^{i}$) and the $i^{th}$ reference summary $y^i\mathbin{:}\{y_{1}^{i},y_{2}^{i},...y_{m}^{i}\}$ consists of $m$ tokens ($j^{th}$ token represented by $y_{j}^{i}$ \& $m<<n$). The standard way to fine-tune $\pi_{ref}$ on $T$ is to simply fine-tune $\pi_{ref}$ using the cross-entropy loss over the original training dataset $D$, as shown in Figure \ref{fig:main}(a). 
% The model fine-tuned using this approach is represented by $\pi_{sft}$.

Aligning $\pi_{ref}$ using alignment training requires the need for preference-based data $D_{pref}\mathbin{:}\{X,Y_{w},Y_{l}\}$, where $Y_{w}$ is a set of preferred summaries, and $Y_{l}$ are the dispreferred ones. Such preference-based data is usually gathered through human annotation or is generated synthetically. As previously explained, not only gathering human annotations is expensive in the clinical domain, but even generating synthetic data using standard approaches like corruption \cite{chen2023purr} can be challenging. So in this work, we \textbf{(1)} propose a new pipeline to generate high-quality synthetic preference data $D_{pref}\mathbin{:}\{X,Y_{w},Y_{l}\}$; \textbf{(2)} use $D_{pref}$ to align $\pi_{ref}$ to generate factually consistent outputs using alignment methods like DPO training \& SALT loss. In the following subsections, we describe the synthetic edit-based preference data generation pipeline and the edit feedback-based alignment training method in detail.

\begin{algorithm}

\footnotesize
    \caption{Synthetic preference data generation (\texttt{\textbf{High$\rightarrow$Low}} \& \texttt{\textbf{Low$\rightarrow$High}}).}
    \DontPrintSemicolon
    \textbf{Dataset:} $D$:$\{X,Y\}$ \\
    \textbf{Clinical Note Articles:} $X$:$\{x^1,...,x^c\}$ \\
    \textbf{Reference Summaries:} $Y$:$\{y^1,...,y^c\}$ \\
    % \textbf{Reference Model:} $\pi_{ref}$ \\
    \textbf{Small LM:} $\pi_{sm}$ \\
    % \textbf{Aligned LM:} $\pi_{\theta}$ \\ \\
    
    \SetKwFunction{HighToLow}{get_$D_{pref}^{High\rightarrow Low}$}
    \SetKwProg{Fn}{Function}{:}{}
    \Fn{\HighToLow{$D$,$f_e^{High\rightarrow Low}$}}{
        $D_{pref}: \{\}$\Comment{\textcolor{magenta}{Preference Data}}\\
        \For{$i=1$ \KwTo $c$}
        {
            $y_{-}^i,E^i\leftarrow f_e^{High\rightarrow Low}(x^i,y^i)$\\
            $y_{w}^i\leftarrow y^i$\Comment{\textcolor{green}{Preferred}}\\
            $y_{l}^i\leftarrow y_{-}^i$\Comment{\textcolor{red}{dispreferred}}\\
            $D_{pref}\leftarrow D_{pref}+\{x^i,y_{w}^i,y_{l}^i\}$
        }
        \textbf{return} $D_{pref}$
    }

    \SetKwFunction{LowToHigh}{get_$D_{pref}^{Low\rightarrow High}$}
    \SetKwProg{Fn}{Function}{:}{}
    \Fn{\LowToHigh{$D$,$f_e^{Low\rightarrow High}$,$\pi_{sm}$}}{
        $D_{pref}: \{\}$\Comment{\textcolor{magenta}{Preference Data}}\\
        \For{$i=1$ \KwTo $c$}
        {
            $y_{*}^i\leftarrow \pi_{small}(x^i)$\Comment{\textcolor{orange}{$\pi_{sm}$ Unaligned Output}}\\
            $y_{+}^i,E^i\leftarrow f_e^{Low\rightarrow High}(x^i,y_{*}^i)$\\
            $y_{w}^i\leftarrow y_{+}^i$\Comment{\textcolor{green}{Preferred}}\\
            $y_{l}^i\leftarrow y_{*}^i$\Comment{\textcolor{red}{dispreferred}}\\
            $D_{pref}\leftarrow D_{pref}+\{x^i,y_{w}^i,y_{l}^i\}$
        }
        \textbf{return} $D_{pref}$
    }
    
    % \emph{Training} \\
\label{alg:sythetic_data_generation}
\end{algorithm}
% \vspace{-4mm}

\section{Synthetic Imitation Edit Feedback} \label{feedback_description}

For summarization alignment, the model learns from the preference data pairs ($Y_{w}$,$Y_{l})$) in $D_{pref}$ by learning to increase the likelihood of $Y_{w}$ and to decrease the likelihood of $Y_{l}$. Usually, $Y_{w}$ is easy to get from the ground truth labels (reference summaries) in $D$, but on the other hand, $Y_{l}$ is usually not readily available. Heuristic-based data augmentation functions have been previously explored to tackle this problem in low-resource settings \cite{kryscinski2019evaluating}. In this work, we propose to use LLMs as synthetic experts (specifically acting as edit functions ($f_e$) imitating as domain experts) to synthetically generate $D_{pref}\mathbin{:}\{X,Y_{w},Y_{l}\}$ in two directions, i.e., (1) \texttt{\textbf{High$\rightarrow$Low}} or (2) \texttt{\textbf{Low$\rightarrow$High}}. The synthetic preference data generation procedure for the above-mentioned two directions is explained in detail in the following two sections.
% to generate $Y_{l}$-$Y_{w}$ pairs have been explored to generate the corrupted summaries, which can be used to act as dispreferred summaries. However leveraging synthetic preference data generated using heuristic-based data augmentation functions for factuality alignment can be challenging and misaligned with the final factuality alignment objective, especially in the clinical domain where some phrases (clinical instructions) in the discharge summaries are extremely important for the correct diagnosis of the patient. So using incorrectly (corrupting incorrect phrases in the summaries of $Y_{+}$) corrupted $Y_{+}$ to act as $Y_{-}$ for aligning the model using DPO can lead to poor factuality alignment.

\subsection{\texttt{High→Low} Synthetic Preference Data Generation} \label{sec:H2LData}
% \setlength\intextsep{1pt}
% \begin{wraptable}{r}{7.5cm}

% \begin{table}[]
% \centering
% \small
% \resizebox{0.8\columnwidth}{!}{%
% \begin{tabular}{cl}
% \hline
% % \multicolumn{3}{c}{Instruction Category Labeling Guideline} \\ \hline
% \multicolumn{1}{c|}{\begin{tabular}[c]{@{}c@{}}Edit\\ Operation\end{tabular}} &                \multicolumn{1}{c}{Description} \\ \hline \hline
% \multicolumn{1}{c|}{\textbf{ADD Operation}} & \begin{tabular}[l]{@{}l@{}}Intentionally \textbf{including} medico-legally phrases in\\ the edited summary from the article or reference\\ summary that \textbf{are required} for accurate\\ diagnosis and treatment documentation.\end{tabular} \\ \cline{2-2}
% \multicolumn{1}{c|}{\textbf{OMIT Operation}} & \begin{tabular}[l]{@{}l@{}} Intentionally \textbf{not including} medico-legally phrases\\ in the edited summary from the article or reference\\ summary that \textbf{are not required} for accurate diagnosis\\ and treatment documentation.\end{tabular} \\ \hline 
% \end{tabular}
% }
% \caption{Factuality Edit Operations}
% \label{tab:factuality_edit_ops}
% % \end{wraptable}
% \end{table}
For the \texttt{\textbf{$\text{High}\rightarrow \text{Low}$}}, to generate high-quality synthetic edit-based preference data for factuality alignment, we use off-the-shelf LLMs like \texttt{GPT-3.5} \& \texttt{GPT-4} to act as synthetic domain experts specifically acting as edit function $f_e^{High\rightarrow Low}$ to mirror edits made by actual domain experts, to generate imitation edit data $Y_{-}$ by adding hallucination to $Y$ (as shown in get\_$D_{pref}^{High\rightarrow Low}()$ synthetic edit data generation function in Algorithm \ref{alg:sythetic_data_generation}). For the \texttt{\textbf{$\text{High}\rightarrow \text{Low}$}} preference data, since $Y$ is the original ground truth in $D$, we treat it as the $Y_{w}$, whereas $Y_{-}$ is treated as $Y_{l}$ as it is the hallucinated summary w.r.t $Y$. In $f_e^{High\rightarrow Low}\mathbin{:}\{x^i,y^i\}\rightarrow y_{-}^i$, we prompt synthetic experts to generate a hallucinated summary given a clinical note $x^i$ and the corresponding reference summary $y^i$, as shown in Figure \ref{fig:main}(c). The prompt for $f_e^{High\rightarrow Low}$ is designed to generate $y_{-}^i$ using edits introduced through the edit operations listed in Table \ref{tab:edit_ops} of Appendix \ref{appendixA}, and the resulting $y_{-}^i$ sounds plausible but includes hallucinated information that is not required for accurate diagnosis and treatment documentation of $x^i$. The detailed prompt is attached in Table \ref{tab:hallucination-prompt} of Appendix \ref{appendixA}. 
% \begin{itemize}
%     \item \textbf{ADD Operation}: Intentionally add medico-legally essential words from the article required for accurate diagnosis and treatment documentation.
%     \item \textbf{OMIT Operation}: Intentionally omit medico-legally essential words in the reference summary required for accurate diagnosis and treatment documentation.
% \end{itemize}

\begin{algorithm*}
\footnotesize
    \caption{DPO \& SALT loss functions for alignment training.}
    \DontPrintSemicolon
    \textbf{Preference Dataset:} $D_{pref}$:$\{X,Y_w,Y_l\}$, where $Y_w$ \& $Y_l$ are generated using get\_$D_{pref}^{Low\rightarrow High}$ or get\_$D_{pref}^{High\rightarrow Low}$ from Algorithm \ref{alg:sythetic_data_generation}.\\
    \textbf{Reference Model:} $\pi_{ref}$ \\
    \textbf{Aligned Model:} $\pi_{\theta}$ \\
    $\ell_{dpo}(\pi_{\theta};\pi_{ref})=-\mathbin{E}_{(x^i,y_{w}^i,y_{l}^i)\sim D_{\text{pref}}}\left(\log \sigma\left[\beta\log\frac{\pi_{\theta}(y_{w}^i|x^i)}{\pi_{ref}(y_{w}^i|x^i)}-\beta\log\frac{\pi_{\theta}(y_{l}^i|x^i)}{\pi_{ref}(y_{l}^i|x^i)}\right]\right)$\Comment{\textcolor{magenta}{DPO Loss}}\\
    $\ell_{\text{salt}}(\pi_{\theta};\pi_{\text{ref}}) = -\mathbb{E}_{(x^i,y_{w}^i,y_{l}^i)\sim D_{\text{pref}}}\left(\alpha_1\Omega_{1} + \alpha_2\Omega_{2} - \alpha_3\Omega_{3}\right)$\Comment{\textcolor{magenta}{SALT Loss}}\\
    $\Omega_{1}=\sum_{a \in A} \log \pi_{\theta}(a|x^i)$\Comment{\textcolor{magenta}{For aligned tokens}}\\
    $\Omega_{2}=\sum_{u_w \in U_{y_{w}^i}} \log \pi_{\theta}(u_w|x^i)$\Comment{\textcolor{magenta}{For unaligned tokens in $y_w^i$}}\\
    $\Omega_{3}=\sum_{u_l \in U_{y_{l}^i}} \log(1 - \pi_{\theta}(u_l|x^i))$\Comment{\textcolor{magenta}{For unaligned tokens in $y_l^i$}}
    
\label{alg:alignment_algo}
\end{algorithm*}

\subsection{\texttt{Low→High} Synthetic Preference Data Generation}
\label{sec:L2HData}
Generation using smaller LMs like \texttt{GPT-2} (even after fine-tuning) has been observed to generate hallucinations and factually incorrect text \cite{zhang2023siren}. By leveraging this phenomenon and treating the summaries generated from smaller LMs (represented by $Y_{*}$) as factually unaligned, we propose an alternative direction (\texttt{\textbf{$\text{Low}\rightarrow \text{High}$}}) to generate high-quality synthetic edit-based preference data for factuality alignment where we use off-the-shelf LLMs (\texttt{GPT-3.5} \& \texttt{GPT-4}) to act as synthetic domain experts specifically acting as edit function $f_e^{Low\rightarrow High}$ to generate imitation edit data $Y_{+}$ by improving factuality in $Y_{*}$ (as shown in get\_$D_{pref}^{Low\rightarrow High}$ synthetic edit data generation function in Algorithm \ref{alg:sythetic_data_generation}). Here for \texttt{\textbf{$\text{Low}\rightarrow \text{High}$}} preference data, since $Y{*}$ is generated from an unaligned smaller model (susceptible to hallucinations and poor generation), we treat it as $Y_{l}$, whereas $Y_{+}$ is treated as $Y_{w}$ as it is the factually improved summary w.r.t $Y_{*}$. In $f_e^{Low\rightarrow High}\mathbin{:}\{x^i,y_{*}^i\}\rightarrow y_{+}^i$, we prompt synthetic experts to generate a factually improved summary given a clinical note $x^i$ and the corresponding smaller model generated summary $y_{*}^i$, as shown in Figure \ref{fig:main}(b). The prompt for $f_e^{Low\rightarrow High}$ is designed to generate $y_{+}^i$ using edits introduced through the edit operations listed in \textbf{Table \ref{tab:edit_ops}} of Appendix \ref{appendixA}, and the resulting $y_{+}^i$ is factually consistent and includes information that is required for accurate diagnosis and treatment documentation of $x^i$. The detailed prompt is attached in Table \ref{tab:factuality-prompt} of Appendix \ref{appendixA}.

Similar to chain-of-thought phenomenon \cite{chu2023survey}, for both $f_e^{High\rightarrow Low}\mathbin{:}\{x^i,y^i\}\rightarrow y_{-}^i$ (\texttt{\textbf{$\text{High}\rightarrow \text{Low}$}}) \& $f_e^{Low\rightarrow High}\mathbin{:}\{x^i,y_{*}^i\}\rightarrow y_{+}^i$ (\texttt{\textbf{$\text{Low}\rightarrow \text{High}$}}), we prompt the synthetic experts to first generate a set of $I$ edit instructions $E^i\mathbin{:}\{e_{1}^i,e_{2}^i,...e_{I}^i\}$, where each instruction consists of either an ADD or OMIT operation (Table \ref{tab:edit_ops} of Appendix \ref{appendixA}; we also provide justification for only using ADD \& OMIT operations for our edits in Appendix \ref{appendixA}) to be done on the contents $X^i$ or $Y^i/Y_{*}^i$. Then the prompt further leverages $E^i$ to generate $y_{-}^i/y_{+}^i$. In summary, we prompt the synthetic expert to ADD or OMIT medico-legally unimportant/important factual information resulting in a decrease/increase in the factual consistency of the contents of $y_{-}^i/y_{+}^i$ respectively. Examples of the generated edit instructions $E^i$ along with the edited summaries $y_{-}^i/y_{+}^i$ are attached in the Appendix \ref{appendixD}.

\subsection{Factual Alignment with Edit Feedback}
\label{sec:alignment_training}
After collecting preference data for imitating edit feedback, we naturally obtain a pair of summaries (low-quality dispreferred and high-quality preferred). For SFT, since it aims to maximize the probability of the model generating certain token distributions, it is evident that only high-quality summaries can be used for optimization.
In contrast, preference training utilizes both low and high-quality summaries from the dataset to align the model toward the desired direction based on their differences.
Specifically, in this paper, we employ two alignment algorithms, DPO and SALT, to align $\pi_{ref}$ based on the differences in factuality levels between the high and low-quality summaries.

\subsubsection{DPO Training}

% \begin{dmath}
% \label{equ:dpo}
% \ell_{dpo}(\pi_{\theta};\pi_{ref})=-\mathbin{E}_{(x^i,y_{w}^i,y_{l}^i)\sim D_{pref}}\left(\log \sigma\left[\beta\log\frac{\pi_{\theta}(y_{w}^i|x^i)}{\pi_{ref}(y_{w}^i|x^i)}-\beta\log\frac{\pi_{\theta}(y_{l}^i|x^i)}{\pi_{ref}(y_{l}^i|x^i)}\right]\right)
% \end{dmath}

For aligning $\pi_{ref}$ using DPO ($\pi_{ref}\rightarrow\pi_{\theta}$), we train the model by optimizing the loss function $\ell_{dpo}$ shown in Algorithm \ref{alg:alignment_algo}, where given the preference data $D_{pref}\mathbin{:}\{X,Y_{w},Y_{l}\}$ consisting of a set of clinical notes $x^i$, preferred summaries $Y_{w}^i$, and the dispreferred summaries $Y_{l}^i$, the model learns to increase the likelihood of $Y_{w}^i$ and to decrease the likelihood of the $Y_{l}^i$. In the equation, $\pi_{ref}$ is the base model and $\pi_{\theta}$ is the model being trained to have improved alignment and $\beta$ is used to scale the weight on how incorrect the model should treat $y_{l}^i$ relative to $y_{w}^i$. The higher the $\beta$ beta, the less the divergence from $\pi_{ref}$.

\subsubsection{SALT Training}

For situations where the differences between the preferred and dispreferred summaries are minimal, the DPO training method may not be optimal, as it tends to reward many tokens in $y_{w}^i$ while penalizing the same tokens in $y_{l}^i$. To address this, we explore the SALT~\cite{yao2023improving} training method in Algorithm \ref{alg:alignment_algo}, specifically designed for feedback involving minor edits. SALT training requires the sequence alignment of $y_{w}^i$ and $y_{l}^i$, identifying:

% If a data does not have a large number of edits (that is, the gap between the two sequences $y_{w}^i$ and $y_{l}^i$ is very small), DPO loss will reward a large number of tokens in $y_{w}^i$ While penalizing the same tokens in $y_{l}^i$, this is not the most appropriate alignment algorithm for edit feedback. In this article we also try another alignment algorithm designed for edit feedback, SALT. It requires sequence alignment of $y_{w}^i$ and $y_{l}^i$ first. We define:

\begin{itemize}[itemsep=0pt, parsep=0pt, topsep=0pt]
    \item $\Omega_{1}$: The set of tokens in $y_{w}^i$ and $y_{l}^i$ that are aligned through sequence alignment, indicating similarity or identical parts between the preferred and dispreferred summaries.
    \item $\Omega_{2}$: The set of tokens in the preferred summary $y_{w}^i$ that cannot be matched with any tokens in the dispreferred summary $y_{l}^i$, representing unique or important aspects of the preferred summary.
    \item $\Omega_{3}$: The set of tokens in the dispreferred summary $y_{l}^i$ that cannot be matched with any tokens in the preferred summary $y_{w}^i$, representing aspects to be avoided or corrected.
\end{itemize}

\noindent Then we can calculate SALT loss as shown in Algorithm \ref{alg:alignment_algo}.
Here, the SALT loss is calculated to enhance model alignment by optimizing the following objectives: promoting the likelihood of tokens in $\Omega_{1}$ and $\Omega_{2}$ while discouraging the likelihood of tokens in $\Omega_{3}$.
The model, denoted as $\pi_{\theta}$, aims to align more closely with the refined preferences represented by the preferred summaries. Weights $\alpha_1$, $\alpha_2$, and $\alpha_3$ adjust the significance of aligned tokens, unique preferred summary tokens, and unique dispreferred summary tokens, respectively, in the loss function. This method allows for a nuanced adjustment of the model's predictions, ensuring that minor but critical edits are appropriately incorporated into the training process.

% {
% \scriptsize
% \begin{dmath}
% \ell_{\text{salt}}(\pi_{\theta};\pi_{\text{ref}}) = -\mathbb{E}_{(x^i,y_{w}^i,y_{l}^i)\sim D_{\text{pref}}}\left(\alpha_1 \sum_{a \in A} \log \pi_{\theta}(a|x^i) 
% + \alpha_2 \sum_{u_w \in U_{y_{w}^i}} \log \pi_{\theta}(u_w|x^i) 
% - \alpha_3 \sum_{u_l \in U_{y_{l}^i}} \log(1 - \pi_{\theta}(u_l|x^i))\right)
% \end{dmath}
% }

% Where:
% \begin{itemize}[itemsep=0pt, parsep=0pt, topsep=0pt]
%     \item $x^i$: The input to the model, such as a clinical note or document.
%     \item $y_{w}^i$: The preferred summary or outcome for input $x^i$.
%     \item $y_{l}^i$: The dispreferred summary or outcome for input $x^i$.
%     \item $D_{\text{pref}}$: The dataset containing triplets of inputs, preferred summaries, and dispreferred summaries, sampled from the preference data.
%     \item $\alpha_1$, $\alpha_2$, $\alpha_3$: Weights for adjusting the contribution of the loss associated with aligned tokens ($A$), unmatched tokens in the preferred summary ($U_{y_{w}^i}$), and unmatched tokens in the dispreferred summary ($U_{y_{l}^i}$), respectively.
%     \item $\pi_{\theta}(a|x^i)$, $\pi_{\theta}(u_w|x^i)$, $\pi_{\theta}(u_l|x^i)$: The probabilities assigned by the model $\pi_{\theta}$ to tokens $a$ (aligned), $u_w$ (unmatched in $y_{w}^i$), and $u_l$ (unmatched in $y_{l}^i$), given the input $x^i$.
% \end{itemize}

\begin{table*}[]
% \vspace{-6mm}
\centering
\resizebox{\textwidth}{!}{%
\begin{tabular}{cccc|ccccc}
\hline
\multicolumn{1}{c|}{\begin{tabular}[c]{@{}c@{}}Data Generation\\ Setting\end{tabular}} & \multicolumn{1}{c|}{\begin{tabular}[c]{@{}c@{}}Synthetic\\ Expert\end{tabular}} & \begin{tabular}[c]{@{}c@{}}\% \textbf{ADD}\\ Instructions\\ (Out of Total)\end{tabular} & \begin{tabular}[c]{@{}c@{}}\% \textbf{OMIT}\\ Instructions\\ (Out of Total)\end{tabular} & \begin{tabular}[c]{@{}c@{}}\% of Total \\ \textbf{Hallucination} Instructions\\ (\textbf{ADD \& OMIT})\end{tabular} & \begin{tabular}[c]{@{}c@{}}\%  \textbf{Hallucination}\\ \textbf{ADD} Instructions\\ (Out of Total)\end{tabular} & \begin{tabular}[c]{@{}c@{}}\%  \textbf{Hallucination}\\ \textbf{OMIT} Instructions\\ (Out of Total)\end{tabular} & \begin{tabular}[c]{@{}c@{}}\%  \textbf{Hallucination}\\ \textbf{ADD} Instructions\\ (Out of ADD Ins.)\end{tabular} & \begin{tabular}[c]{@{}c@{}}\%  \textbf{Hallucination}\\ \textbf{OMIT} Instructions\\ (Out of OMIT Ins.)\end{tabular} \\ \hline
\multicolumn{1}{c|}{\multirow{2}{*}{High $\rightarrow$ Low}} & \multicolumn{1}{c|}{\texttt{GPT-4}} & ${52.00}$ & ${48.00}$ & ${54.33}_{23.35}$ & ${18.50}_{16.12}$ & ${34.00}_{7.94}$ & ${35.00}_{30.41}$ & ${71.67}_{16.07}$ \\  \cline{2-2}
\multicolumn{1}{c|}{} & \multicolumn{1}{c|}{\texttt{GPT-3.5}} & ${64.16}$ & ${35.84}$ & ${33.06}_{15.49}$ & ${14.48}_{13.30}$ & ${28.11}_{4.20}$ & ${18.89}_{18.36}$ & ${70.67}_{11.14}$ \\ \hline

& & & & \begin{tabular}[c]{@{}c@{}}\% of Total \\ \textbf{Factuality} Instructions\\ (\textbf{ADD \& OMIT})\end{tabular} & \begin{tabular}[c]{@{}c@{}}\%  \textbf{Factuality}\\ \textbf{ADD} Instructions\\ (Out of Total)\end{tabular} & \begin{tabular}[c]{@{}c@{}}\%  \textbf{Factuality}\\ \textbf{OMIT} Instructions\\ (Out of Total)\end{tabular} & \begin{tabular}[c]{@{}c@{}}\%  \textbf{Factuality}\\ \textbf{ADD} Instructions\\ (Out of ADD Ins.)\end{tabular} & \begin{tabular}[c]{@{}c@{}}\%  \textbf{Factuality}\\ \textbf{OMIT} Instructions\\ (Out of OMIT Ins.)\end{tabular}\\ \hline
\multicolumn{1}{c|}{\multirow{2}{*}{Low $\rightarrow$ High}} & \multicolumn{1}{c|}{\texttt{GPT-4}} & ${54.33}$ & ${45.67}$ & ${42.59}_{17.14}$ & ${28.67}_{15.74}$ & ${4.39}_{1.29}$ & ${51.67}_{28.87}$ & ${11.67}_{2.89}$ \\  \cline{2-2}
\multicolumn{1}{c|}{} & \multicolumn{1}{c|}{\texttt{GPT-3.5}} & ${55.00}$ & ${45.00}$ & ${28.64}_{3.11}$ & ${24.50}_{2.36}$ & ${4.14}_{1.47}$ & ${47.22}_{4.81}$ & ${7.50}_{2.50}$\\ \hline
\end{tabular}%
}
% \vspace{-2mm}
\caption{ \textbf{[Columns 3 \& 4]:} Statistics for \% of ADD \& OMIT instructions present in $E^i \in D_{eval}$. \textbf{[Column 5]:} Statistics for \% of both ADD \& OMIT instructions present in $E^i \in D_{eval}$ that were annotated as hallucinating (up)/factuality aid (down) instruction. \textbf{[Column 6 \& 7]:} Statistics for \% of only ADD or OMIT instructions respectively out of total (ADD + OMIT) instructions present in $E^i \in D_{eval}$ that were annotated as hallucinating (up)/factuality aid (down) instruction. \textbf{[Column 8 \& 9]:} Statistics for \% of only ADD or OMIT instructions respectively out of respective ADD/OMIT instructions present in $E^i \in D_{eval}$ that were annotated as hallucinating (up)/factuality aid (down) instruction.}
% \vspace{-6mm}
\label{tab:edit_analysis}
\end{table*}

\section{Results}
\subsection{Experimental Setup}
In the next section, we evaluate the quality of our synthetic edit-based preference data (edit instructions $E^i$ as well as the edited summaries $y_{-}^i$ \& $y_{+}^i$) generated using our synthetic experts in our pipeline for both the directions: \texttt{\textbf{$\text{High}\rightarrow \text{Low}$}} \& \texttt{\textbf{$\text{Low}\rightarrow \text{High}$}}. For our pipeline, we experimented with both (\texttt{GPT-3.5} \& \texttt{GPT-4}) as our synthetic experts. We leverage human annotations from domain expert human annotators on a small sample\footnote{For our human evaluation, we used 10 samples from our synthetic data using three domain experts: 1 doctor \& 2 medical students. Human evaluation guideline: Appendix \ref{appendixB}.} of our pipeline-generated edits (called human evaluation sample set $D_{eval}$) to quantify these results. We also conduct experiments for external evaluation of our synthetic edit-based preference data on the downstream clinical note summarization task. Following previous works \cite{cai-etal-2022-generation,adams-etal-2022-learning} in this domain, we used their cleaned discharge instruction dataset which is based on \texttt{MIMIC-III} database \cite{johnson2016mimic} (instead of \texttt{MIMIC-IV}) in our experiments for clinical note summarization. This dataset consists of 25k/3k/3k train/valid/test respective clinical notes and reference summaries. Due to resource limitations, we restricted the downstream task train/valid/test set to 5k/128/128 samples, whereas for \texttt{\textbf{Low$\rightarrow$High}}, we used the held-out training set of 20k samples to fine-tune $\pi_{sm}$.

In Section \ref{sec:external_evaluation}, we further evaluate the effectiveness of our generated synthetic preference data for improving factuality in the weaker LLM-generated outputs. For showcasing extrapolation of our approach to different model parameter scales, in our experiments, we use \texttt{GPT-2 (1.5B)} \& \texttt{Llama-2 (7B)} models (hyperparameter details in Appendix \ref{appendixf}). Although \texttt{GPT-2} is an old model and now there exist larger and better models like Llama variants, it is still important to showcase the applicability of our approach on smaller and weaker models as they are very frequently used on device deployed models, discussed more in Section \ref{sec:ethics}. We compare the summarization performance of the models trained using a simple STF approach vs the preference-based DPO/SALT training approach. Human evaluation examples are listed in Appendix \ref{appendixE}.

\subsection{Synthetic Edit Feedback Evaluation} \label{section4_1}

To quantify the quality of the edit instructions generated by our proposed pipeline, we used domain expert annotators to annotate the generated edit instructions and edited summaries in $D_{eval}$.
In our proposed pipeline, both $f_e^{High\rightarrow Low}$ \& $f_e^{Low\rightarrow High}$, first generates a set of edit instruction $E^i$ which are then used to generate $y_{-}^i$ or $y_{+}^i$ respectively. 
To quantify the quality of $y_{-}^i$/$y_{+}^i$, we first used human evaluation to evaluate the quality of the corresponding $E^i$ by annotating whether a generated instruction $e_{j}^i$ in $E^i$ is useful or not in generating hallucinations (in the case of \texttt{\textbf{$\text{High}\rightarrow \text{Low}$}})/factuality improvements (in the case of \texttt{\textbf{$\text{Low}\rightarrow \text{High}$}}).
Refer to Table \ref{tab:edit_analysis} for human evaluation results on generated edit instructions. 
The numbers in Table \ref{tab:edit_analysis} over the last five columns report the mean and standard deviation of the hallucination/factuality statistics over annotations by all our annotators, whereas columns 3 \& 4 are the statistics derived directly from $E^i \in D_{eval}$.
For \texttt{\textbf{High$\rightarrow$Low}} setting, we ask the annotators to annotate whether the edit instruction is hallucinating instructions or not, whereas for \texttt{\textbf{Low$\rightarrow$High}} setting, we ask the annotators to annotate if the edit instructions are factuality aid instructions or not.
Here are the results from our human evaluation of synthetic edit data generation:

\begin{figure}
% \vspace{-6mm}
    \begin{center}
    \includegraphics[width=1\columnwidth]{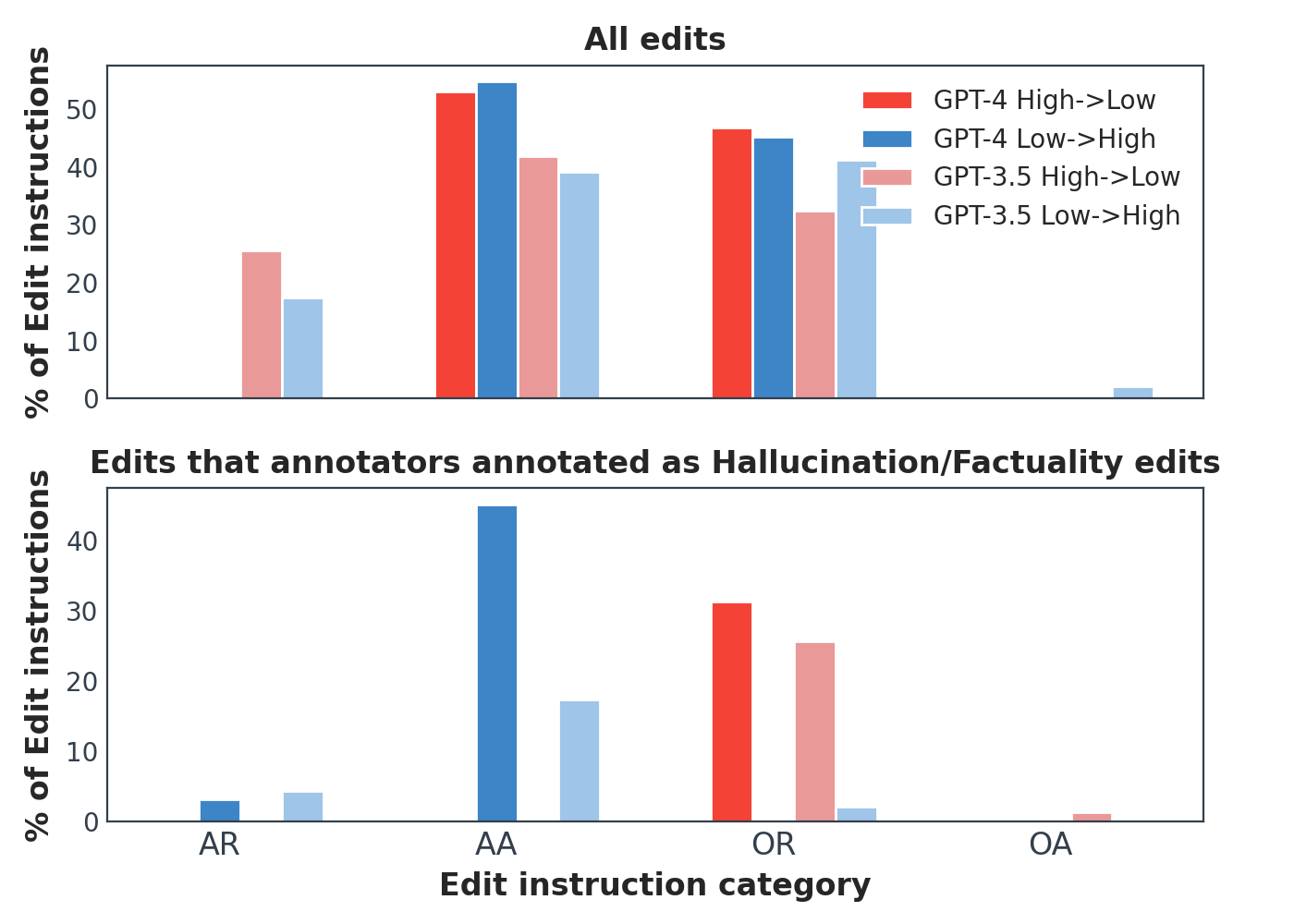}
    \end{center}
    % \vspace{-5mm}
    \caption{\textbf{Top:} \% of edits made in the $D_{eval}$ for each edit type listed in Table \ref{tab:edit_types} of Appendix \ref{appendixc}. \textbf{Bottom:} \% of edits that resulted in hallucinations/factuality aid according to the annotators. Plot legend format: \texttt{<Synthetic Expert> <Data Generation Setting>}}
    \label{fig:edit_category_analysis}
% \end{wrapfigure}
% \vspace{-5mm}
\end{figure}

\paragraph{\texttt{GPT-4} is much better at following prompt instructions} We designed our prompt to use equal number of ADD \& OMIT hallucination (\texttt{\textbf{High$\rightarrow$Low}})/factuality aid (\texttt{\textbf{Low$\rightarrow$High}}) edit instructions, but we observe that \texttt{GPT-3.5} is poor at following these instructions relative to \texttt{GPT-4}, which in both $f_e^{High\rightarrow Low}$ \& $f_e^{Low\rightarrow High}$ generates approximately equal number of ADD \& OMIT instructions. From Table \ref{tab:edit_analysis} we also observe that \texttt{GPT-4} is better at generating a higher percentage of desired edit instructions (last two columns of Table \ref{tab:edit_analysis}) which according to our annotators actually leads to desired edits. Using \texttt{GPT-4} for edits also resulted in a higher percentage of the total number of desired edits generated relative to \texttt{GPT-3.5} (\textbf{Column 5}), suggesting better prompt following tendencies.

\paragraph{OMIT instructions leads to hallucination whereas ADD instructions lead to factuality improvements} We observe that in the case of \textbf{High$\rightarrow$Low}, majority of actual hallucination edits are generated using OMIT instructions (\textbf{Column 9 [Up]}), whereas in the case of \texttt{\textbf{Low$\rightarrow$High}}, majority of actual factuality edits are generated using ADD instructions (\textbf{Column 8 [Down]}). While the percentage of ADD \& OMIT edits are relatively higher in the case of \texttt{\textbf{Low$\rightarrow$High}} \& \texttt{\textbf{High$\rightarrow$Low}} respectively, in the case of \texttt{\textbf{Low$\rightarrow$High}} it is relatively much more skewed towards the ADD instructions leading to highly skewed edit type distribution in $E^i$ from $f_e^{Low\rightarrow High}$. Counter-intuitively, this suggests that for our synthetic experts, it is relatively difficult to identify and remove factually incorrect information compared to factually correct information. To quantify the type of edits made in our pipeline, we prompted (\ref{appendixc}) \texttt{GPT4} to categorize edit instructions into one of the edit-type categories listed in Table \ref{tab:edit_types} of Appendix \ref{appendixc}. From the results in Figure \ref{fig:edit_category_analysis} for \texttt{\textbf{High$\rightarrow$Low}} setting, the majority of desired hallucination edits are generated by either omitting useful information from the reference summary/article, whereas in the case of \texttt{\textbf{Low$\rightarrow$High}}, majority of desired factuality improvement edits are generated adding useful information from the article/unaligned model generated summary. This further validates our findings.

% \begin{figure}
%     \begin{center}
%     \includegraphics[width=0.9\columnwidth]{Images/bar_plot_with_error_bars.png}
%     \end{center}
%     \caption{Annotator edit summary preference analysis.}
%     \label{fig:preference}
% % \end{wrapfigure}
% \end{figure}

\begin{table}%[h]
% \vspace{-6mm}
\centering
\resizebox{1\columnwidth}{!}{%
\begin{tabular}{c|cccc}
\hline
\begin{tabular}[c]{@{}l@{}}Edit Data\\ Setting\end{tabular} & \multicolumn{2}{c|}{\texttt{High$\rightarrow$ Low}}            & \multicolumn{2}{c}{\texttt{Low$\rightarrow$ High}} \\ \hline
Prompt Model                                             & \multicolumn{1}{c|}{\texttt{GPT-4}} & \multicolumn{1}{c|}{\texttt{GPT-3.5}} & \multicolumn{1}{c|}{\texttt{GPT-4}}      & \texttt{GPT-3.5}     \\ \hline
Preference \%  &  $6.67_{0.05}$  & $6.67_{0.12}$ &  ${\textbf{26.67}}_{\textbf{0.15}}$ &  ${\textbf{33.33}}_{\textbf{0.15}}$ \\ \hline
\end{tabular}%
}
% \vspace{-4mm}
\caption{Mean \& std. statistics for annotated preference percentage of (1) \textbf{Column [2 \& 3]:} $y_{-}^i$ over $y^i$, and (1) \textbf{Column [4 \& 5]:} $y_{+}^i$ over $y_{*}^i$. $\downarrow$ pref. \% in \texttt{\textbf{High$\rightarrow$Low}} \& $\uparrow$ pref. \% in \texttt{\textbf{Low$\rightarrow$High}} is good.}
% \vspace{-4mm}
\label{tab:preference}
\end{table}

\noindent\textbf{\textbf{High$\rightarrow$Low} synthetic data results in better quality preference data compared to \texttt{\textbf{Low$\rightarrow$High}}} From Table \ref{tab:preference}, we observe that for \texttt{\textbf{High$\rightarrow$Low}} the annotators had a very low preference in terms of factuality (<$10$\%,) towards the edited summary, validating our desired output. On the other hand, for \texttt{\textbf{Low$\rightarrow$High}} a relatively higher preference towards the edited summary was observed. Although this is in line with the desired outcome, in case of \texttt{\textbf{Low$\rightarrow$High}}, despite a significant relative increase in preference rate towards the edited summaries (hypothesized to be factually improved), the annotators still preferred the low-quality \(\pi_{sm}\) generated summary on average, suggesting a relatively poor desired preference data.

\noindent\textbf{\texttt{GPT-4} exhibits tendencies to generate higher granularity of synthetic edits relative to \texttt{GPT-3.5}} We further analyzed the edits by quantifying the granularity of edits using \texttt{ROUGE} scores. In Table \ref{tab:edit-granuality}, we calculated \texttt{ROUGE-1/2/L/Lsumm} between the $y^i$ \& $y_{-}^{i}$ in the case of \texttt{\textbf{High$\rightarrow$Low}} preference data, and between the $y_{*}^i$ \& $y_{+}^{i}$ in the case of \texttt{\textbf{Low$\rightarrow$High}} to measure the token level difference between the pair of summaries. From our results we observe that for both directions, \texttt{GPT-4} had significantly higher \texttt{ROUGE} scores relative to \texttt{GPT-3.5}, suggesting \texttt{GPT-4} edits are of high granularity not only at the token level but also for longer spans of tokens.

\subsection{External Evaluation}
\label{sec:external_evaluation}

To evaluate the effectiveness of our proposed edits for improving factuality in the model-generated outputs, we compare the summarization performance of the model trained using a simple STF approach vs the preference-based DPO/SALT training approach, where we use either \texttt{\textbf{High$\rightarrow$Low}} or \texttt{\textbf{Low$\rightarrow$High}} edit pipeline for generating preference data.
In SFT, the model takes a clinical note as input and aims to generate a summary that matches the reference as closely as possible. 
In preference training, using the DPO approach, the goal remains similar but with a focus on favoring a high-quality summary over a low-quality one. 
For SALT, we first target \texttt{GPT}-verified tokens ($\Omega_{1}$), \texttt{GPT}-preferred tokens ($\Omega_{2}$), and \texttt{GPT}-dispreferred tokens ($\Omega_{3}$) using sequence alignment, and then the objective is to increase the likelihood of $\Omega_{1}$ and $\Omega_{2}$ (can be different weights) while decreasing the likelihood of $\Omega_{3}$.
We experiment with \texttt{GPT-2} and \texttt{Llama2} and evaluate the quality of the trained models for summarization using \texttt{ROUGEL} and for factuality using \texttt{G-Eval} and \texttt{UMLS-F1}.
\texttt{G-Eval} evaluates factual alignment using the \texttt{GPT-4} chain-of-thought to assess the factuality when prompted to generate a factuality score (Appendix \ref{appendixD}), whereas \texttt{UMLS-F1} calculates the F1-score between the UMLS medical terms present in the reference summary and the generated summary. For both \texttt{G-Eval} and \texttt{UMLS-F1}, the higher the score, the higher the factuality in the generated output.
We also had two medical students review 50 summaries for factual accuracy, specifically looking for missing or incorrect information that could lead to errors in medical treatment after discharge. 
Each method listed in Table~\ref{table:HIL_results} and~\ref{table:L2H_results} was evaluated in a head-to-head comparison against its corresponding SFT baseline, with the stipulation that no ties were allowed in the assessment.

\begin{table}%[h]
% \vspace{-6mm}
\centering
\resizebox{0.9\columnwidth}{!}{%
\begin{tabular}{c|c|cccc}
\hline
\begin{tabular}[c]{@{}c@{}}Data Generation\\ Setting\end{tabular} & \begin{tabular}[c]{@{}c@{}}Synthetic\\ Expert\end{tabular} & R-1 & R-2 & R-L & RLsum \\ \hline
\multirow{2}{*}{High--\textgreater Low} & \texttt{GPT-4} & \textbf{0.86} & \textbf{0.82} & \textbf{0.84} & \textbf{0.85} \\ \cline{2-2}
 & \texttt{GPT-3.5} & 0.55 & 0.47 & 0.52 & 0.52 \\ \hline
\multirow{2}{*}{Low--\textgreater{}High} & \texttt{GPT-4} & \textbf{0.73} & \textbf{0.68} & \textbf{0.71} & \textbf{0.71} \\ \cline{2-2}
 & \texttt{GPT-3.5} & 0.55 & 0.50 & 0.50 & 0.53 \\ \hline
\end{tabular}%
}
% \vspace{-3mm}
\caption{Edit granuality analysis using \texttt{ROUGE} between (1) \textbf{Row [2 \& 3]:} $y_{-}^i$ \& $y^i$, and \textbf{Row [4 \& 5]:} $y_{+}^i$ \& $y_{*}^i$. Here, $\uparrow$ \texttt{ROUGE} scores signifies $\uparrow$ granuality edits.}
% \vspace{-4mm}
\label{tab:edit-granuality}
\end{table}

As demonstrated in Table~\ref{table:HIL_results} and~\ref{table:L2H_results}, we observed for both \texttt{GPT-2} and \texttt{Llama2}, preference training with \texttt{GPT-4} edits surpassed \texttt{GPT-3.5} edits in performance.
Specifically, SALT using \texttt{GPT-4} edits excelled in all metrics, while SALT with \texttt{GPT-3.5} edits occasionally fell short compared to DPO with \texttt{GPT-3.5} edits. 
This aligns with the conclusions drawn in Section~\ref{section4_1}, suggesting that SALT benefits from higher granularity edit feedback, enabling better alignment outcomes.
\texttt{GPT-3.5} edits, often involving extensive sentence-level modifications, reduce the accuracy of sequence alignment in SALT, introducing noise that degrades the alignment's effectiveness.
In addition, \texttt{GPT-2}'s performance improved with \texttt{\textbf{High$\rightarrow$Low}} over \texttt{\textbf{Low$\rightarrow$High}} in ROUGEL scores, without noticeable differences in factuality and human assessments. 
We only reported \texttt{\textbf{Low$\rightarrow$High}} outcomes for \texttt{GPT-2} in Table~\ref{table:L2H_results} because \texttt{Llama2}, when trained with \texttt{GPT}-edits from the \texttt{\textbf{Low$\rightarrow$High}} pipeline, lagged behind the SFT baseline significantly (Appendix Table~\ref{table:L2H_llama2_results}). 
This discrepancy could be attributed to the \texttt{\textbf{Low$\rightarrow$High}} data being generated by the smaller LM in our experiments, making the corrections from \texttt{GPT}-edits on low-quality data too simplistic for LLMs like \texttt{Llama 2} to achieve factual alignment. 
Conversely, the \texttt{\textbf{High$\rightarrow$Low}} data, crafted by leveraging \texttt{GPT} based on the reference summary, is model-agnostic, enabling more effective utilization for both \texttt{GPT-2} and \texttt{Llama 2}.

\begin{table}%[h]
% \vspace{-6mm}
\centering
\scalebox{0.64}{
\begin{tabular}{c|cccc}
\hline
\multirow{2}{*}{} & ROUGEL & UMLS-F1 & G-Eval & Human H2H (vs X_{sft})
\\

 & \multicolumn{4}{c}{\texttt{GPT-3.5} Synthetic Dataset}
\\
\hline

\texttt{GPT2}-\small{sft} &  34.51 & 31.83 & 3.96 & -
\\

\texttt{GPT2}-\small{dpo} &  34.70 & 32.82 & 4.24 & 62\% win
\\

\texttt{GPT2}-\small{salt} &  34.34 & 32.03 & 4.04 & 44\% win
\\
\hline

\texttt{Llama2}-\small{sft} &  38.03 & 35.47 & 6.48 & -
\\

\texttt{Llama2}-\small{dpo} &  38.15 & 36.30 & 6.61 & 60\% win
\\

\texttt{Llama2}-\small{salt} &  38.32 & 36.85 & 6.54 & 56\% win
\\

\hline

& \multicolumn{4}{c}{\texttt{GPT-4} Synthetic Dataset}
\\
\hline

\texttt{GPT2}-\small{sft} &  34.51 & 31.83 & 3.96 & -
\\

\texttt{GPT2}-\small{dpo} &  37.35 & 34.76 & 4.42 & 78\% win
\\

\texttt{GPT2}-\small{salt} &  38.55 & 36.47 & 4.60 & 72\% win
\\
\hline

\texttt{Llama2}-\small{sft} &  38.03 & 35.47 & 6.48 & -
\\

\texttt{Llama2}-\small{dpo} &  40.47 & 36.82 & 6.71 & 60\% win
\\

\texttt{Llama2}-\small{salt} &  40.50 & 37.51 & 6.82 & 74\% win
\\

\hline
\end{tabular}
}
% \vspace{-3mm}
\caption{External Evaluation \texttt{\textbf{High$\rightarrow$Low}}. First column represents \texttt{<Weaker LLM>-<Training Algorigthm>}} 

\label{table:HIL_results}
% % \vspace{-6mm}
\end{table}

\section{Related Work}
Recent studies have demonstrated the efficacy of LLMs in enhancing data augmentation processes \cite{li2023feasibility, Dai2023ChatAugLC, zhou2022large, dai2022promptagator}. 
Investigations into the precision and effectiveness of LLMs for data annotation have revealed their potential to match or even exceed the accuracy of human annotators, as reported by \citealp{gilardi2023chatgpt} and \citealp{ding2022gpt}. Moreover, the use of LLMs for generating positive sample pairs, crucial for training downstream models, has been explored with promising outcomes \cite{bonifacio2022inpars}. 
Within the biomedical field, LLMs are increasingly utilized for tasks including clinical text mining, question answering, summarization, medical documentation, and other clinical generation tasks for enhancing data augmentation processes.
These efforts aim to address challenges such as suboptimal performance, adherence to instructions, and privacy concerns, showcasing the broad applicability of LLMs in this domain \cite{tang2023does, tran2023bioinstruct, sarker2023medical, wang2023notechat, liao2023differentiate}.

\begin{table}%[h]
% \vspace{-6mm}
\centering
\scalebox{0.66}{
\begin{tabular}{c|cccc}
\hline
\multirow{2}{*}{} & ROUGEL & UMLS-F1 & G-Eval & Human H2H (vs X_{sft})
\\
 & \multicolumn{4}{c}{\texttt{GPT-3.5} Synthetic Dataset}
\\
\hline
SFT &  34.51 & 31.83 & 3.96 & -
\\
DPO &  34.34 & 32.94 & 4.44 & 66\% win
\\
SALT &  34.55 & 33.33 & 4.12 & 53\% win
\\
\hline
\hline
& \multicolumn{4}{c}{\texttt{GPT-4} Synthetic Dataset}
\\
\hline
SFT &  34.51 & 31.83 & 3.96 & -
\\
DPO &  34.30 & 33.64 & 4.34 & 72\% win
\\
SALT &  36.95 & 34.35 & 4.48 & 74\% win
\\
\hline
\end{tabular}
}
% \vspace{-2mm}
\caption{External evaluation \texttt{\textbf{Low$\rightarrow$High}} results (\texttt{GPT2}).} 

\label{table:L2H_results}
% \vspace{-6mm}
\end{table}

On the other hand, \citealp{10.5555/3495724.3495977} notes that standard sequence-to-sequence training (SFT) incorrectly weighs significant errors (like hallucinations) and minor mistakes (such as grammatical inaccuracies) equally, impacting the ability to consistently generate text of high quality as determined by human standards, such as factuality. Recent studies highlight the potential of learning with human feedback paradigms to produce text that meets these high-quality standards \cite{ziegler2019fine, stiennon2020learning, akyurek2023rl4f, dong2023raft, zhao2023slic, yuan2023rrhf}. In the clinical realm, the risk of factual errors by large language models (LLMs) due to gaps in medical knowledge is significant, potentially leading to severe consequences like misdiagnoses \cite{petroni2019language, sung2021can, yao2022extracting, yao2022context}. Various feedback mechanisms—such as comparison-based, scalar, label, edit, and language feedback—have been explored, with calls for further research into less conventional methods like edit and language feedback \cite{casper2023open}. The use of edit feedback in clinical settings, where doctors review AI-generated summaries, presents a practical method for acquiring expert feedback without compromising privacy \cite{yao2023improving}. The proposal to create a synthetic dataset of imitation edit feedback using models such as \texttt{GPT-3.5} and \texttt{GPT-4} offers a promising solution to address privacy concerns \cite{mishra2023synthetic}. Despite the proficiency of \texttt{GPT} models in numerous clinical NLP tasks, including passing the Medical Licensing Examination \cite{kung2023performance, gilson2023does,Yang2023.10.26.23297629}, their ability to generate expert-level edit feedback for clinical applications has not been thoroughly investigated. This paper aims to fill this research gap by evaluating the capability of these \texttt{GPT} variants to effectively generate such feedback.

\section{Conclusion}
This study leverages synthetic edit feedback to improve factual accuracy in clinical summarization using DPO and SALT techniques. Our approach demonstrates the effectiveness of \texttt{GPT}-generated edits in enhancing the reliability of clinical NLP applications.

% In conclusion, this study introduces an innovative approach to enhancing factual consistency in clinical note summarization by leveraging synthetic edit feedback generated by \texttt{GPT-3.5} and \texttt{GPT-4}. Through the application of advanced alignment algorithms such as DPO and SALT, we demonstrate significant improvements in factual accuracy and ROUGEL scores for both GPT-2 and LLaMA-2 models. Our findings highlight the potential of using \texttt{GPT}-generated edits for refining LMs and LLMs in clinical NLP tasks, offering a promising direction for future research in generating more reliable and factually consistent medical summaries.

\section{Limitations and Ethical Considerations}
\label{sec:ethics}

This study offers valuable insights but also comes with several limitations that we would like to highlight:

\begin{itemize}
    \item \textbf{Domain Specificity:} Our research exclusively focuses on the task of factuality alignment in clinical summarization. The adaptation of the proposed method to other domains remains unexplored. This suggests that our approach may need further validation and adjustments before being applied to different fields.
    \item \textbf{Expertise of Annotators:} We relied on 1 doctor and 2 medical students as annotators for human evaluation and preference results. While they are qualified to read and annotate clinical notes and their corresponding discharge summaries, employing more qualified domain experts as annotators would enhance the statistical significance of our results. We leave this to future work, along with addressing concerns about fairness, generalizability to other domains/languages, and potential biases inherent in LLMs.
\end{itemize}

\paragraph{Privacy Implications}
Privacy protection is crucial when dealing with clinical text and patient data. Even though we utilized a de-identified public dataset (such as MIMIC-III), generating and using synthetic data in practical applications must strictly adhere to data protection laws and ethical standards to prevent misuse of patient information.

\paragraph{Bias Considerations}
LLMs may inherently contain or amplify biases present in the training data. When generating edit feedback and synthetic data using LLMs, these biases must be carefully considered to avoid propagating inaccurate or biased information in clinical decision-support tools.

\paragraph{Broader Impacts}
Our research aims to enhance the factual accuracy of clinical summarizations through synthetic edit feedback, potentially positively impacting the reliability of healthcare decision-support systems and reducing patient risk. However, technology usage should be approached cautiously to ensure that technological errors do not endanger patient safety.

\paragraph{Experimentation with more capable LLMs}
We focused our experimentation on only \texttt{GPT-2 (1.5B)} \& \texttt{Llama-2 (7B)} as our weaker LLMs and \texttt{GPT-3.5} \& \texttt{GPT-4} as our synthetic experts, but in future we would also like to explore those capabilities of our model using more recent, capable and domain-specific models like BioGPT and BioLlama, we will leave that to future work.

In summary, while our study demonstrates the potential of using LLMs to improve the factual accuracy of clinical summaries, practical applications must consider domain adaptability, annotator expertise, privacy, bias, and broader ethical societal implications. Future work should focus on addressing these limitations and ethical considerations to ensure the safe, fair, and effective use of technology.

% \newpage

\bibliography{anthology, custom}

\appendix
\section{Edit Prompts} \label{appendixA}
\begin{table}[h]
\centering
\small
\resizebox{1\columnwidth}{!}{%
\begin{tabular}{cl}
\hline
\multicolumn{2}{c}{\textbf{Hallucination Edit Operations}} \\ \hline
\multicolumn{1}{c|}{\begin{tabular}[c]{@{}c@{}}Edit\\ Operation\end{tabular}} &                \multicolumn{1}{c}{Description} \\ \hline
\multicolumn{1}{c|}{\textbf{ADD}} & \begin{tabular}[l]{@{}l@{}}Intentionally \textbf{including} medico-legally phrases in the edited \\summary from the article or reference summary that \textbf{are not} \\\textbf{required} for accurate diagnosis and treatment documentation.\end{tabular} \\ \cline{2-2}
\multicolumn{1}{c|}{\textbf{OMIT}} & \begin{tabular}[l]{@{}l@{}}Intentionally \textbf{not including} medico-legally phrases in the edited\\ summary from the article or reference summary that \textbf{are requi-}\\ \textbf{-red} for accurate diagnosis and treatment documentation.\end{tabular} \\ \hline \hline
\multicolumn{2}{c}{\textbf{Factuality Edit Operations}} \\ \hline
\multicolumn{1}{c|}{\begin{tabular}[c]{@{}c@{}}Edit\\ Operation\end{tabular}} &                \multicolumn{1}{c}{Description} \\ \hline \hline
\multicolumn{1}{c|}{\textbf{ADD}} & \begin{tabular}[l]{@{}l@{}}Intentionally \textbf{including} medico-legally phrases in the edited \\summary from the article or reference summary that \textbf{are required}\\ for accurate diagnosis and treatment documentation.\end{tabular} \\ \cline{2-2}
\multicolumn{1}{c|}{\textbf{OMIT}} & \begin{tabular}[l]{@{}l@{}} Intentionally \textbf{not including} medico-legally phrases in the edited\\ summary from the article or reference summary that \textbf{are not}\\ \textbf{required} for accurate diagnosis and treatment documentation.\end{tabular} \\ \hline
\end{tabular}
}
\caption{Description of edit operations used in our prompts for generation synthetic data in \texttt{\textbf{Hight$\rightarrow$Low}} (hallucination edit operations) \& \texttt{\textbf{Low$\rightarrow$High}} (factuality edit operations)}
\label{tab:edit_ops}
% \vspace{-6mm}
% \end{wraptable}
\end{table}

We used ADD \& OMIT operations as the only to operations available to the synthetic experts to edit and generate new summaries. The definitions for these ADD \& OMIT operations in both \texttt{\textbf{High$\rightarrow$Low}} \& \texttt{\textbf{Low$\rightarrow$High}} settings are given in Table \ref{tab:edit_ops}. The detailed prompts for generating synthetic edit data in both directions, \texttt{\textbf{High$\rightarrow$Low}} \& \texttt{\textbf{Low$\rightarrow$High}} used for generating hallucinations and factuality improvements are provided in Table \ref{tab:hallucination-prompt} \& \ref{tab:factuality-prompt} respectively.

We only include these ADD \& OMIT operations in our prompts because ADD/OMIT is the natural way in which medical professionals create/correct this data. When human experts try to correct AI summary, they can modify or delete a span of tokens, insert a new span of tokens, or not change anything to a span of tokens, all these conditions can be decomposed into ADD \& OMIT combination. Here’s how different editing situations can be decomposed into ADD and OMIT combinations:

\begin{enumerate}
    \item Modify a Span of Tokens:
    \begin{description}
        \item [Decomposition:] First, OMIT the span of tokens that need modification. Then, ADD the new span of tokens with the corrected information.
        \item [Example:] “The patient has a high fever” to “The patient has a mild fever”, first OMIT “high”, then ADD “mild” in its place.
    \end{description}
    \item Delete a Span of Tokens:
    \begin{description}
        \item [Decomposition:] OMIT the span of tokens without adding any new content.
        \item [Example:] To remove “due to viral infection” from “The patient has a mild fever due to viral infection”, simply OMIT “due to viral infection”.
    \end{description}
    \item Insert a New Span of Tokens:
    \begin{description}
        \item [Decomposition:] ADD the new span of tokens at the specific location without omitting any existing content.
        \item [Example:] To add “and coughing” to “The patient has a mild fever”, ADD “and coughing” at the end of the sentence.
    \end{description}
    \item No Change to a Span of Tokens:
    \begin{description}
        \item [Decomposition:] Neither ADD nor OMIT actions are performed on the span of tokens.
        \item [Example:] If “The patient has a mild fever” is accurate and requires no modification, then no action is taken.
    \end{description}
\end{enumerate}
% \usepackage{graphicx}
\begin{table*}[]
\centering
\caption{Hallucination Prompt}
\label{tab:hallucination-prompt}
\resizebox{0.8\textwidth}{!}{%
\begin{tabular}{p{\linewidth}}
\hline
\textbf{»»»» Instruction »»»»} \\
You are a clinical writing assistant who is in edit mode. You are tasked with generating hallucinated summary based on provided a clinical note article and a reference summary for the article. The goal is to edit the reference summary to generate a hallucinated summary that sounds plausible but includes edits introduced through an edit operation which can be one of the following: \\ \\
\textbf{Add Operation:} Intentionally add medico-legally essential words from the article not required for accurate diagnosis and treatment documentation. \\
\textbf{Omit Operation:} Intentionally omit medico-legally essential words in the reference summary required for accurate diagnosis and treatment documentation. \\ \\
For these operations focus on words that, if missing or incorrect in the hallucinated summary, could lead to wrong diagnoses and treatments in the future. Maintain coherence while excluding essential terms. The hallucinated summary should be concise and contain no more than FIVE EXTRA WORDS compared to the reference summary and should have an equal number of Add/Omit operations. \\ \\
Steps for generating the hallucinated summary: \\ \textbf{Step 1:} List the proposed edit operations to introduce hallucination on the reference summary.\\ \textbf{Step 2:} Use the proposed edit operations to edit the reference summary. \\
\textbf{»»»» Output Format »»»»} \\
The output format is: \\
Numbererd List hallucination edits made:\\
\{Edit 1\}, \{Edit 2\}, \{Edit 3\} ...\\
Hallucinated Summary: \\
\textbf{»»»» Follow the above Instructions, Hallucination Method and Output Format »»»»} \\
Now, let's start.\\ Generate the hallucinated summary:\\
Article - \{src\}\\
Reference Summary - \{ref\} \\ \hline
\end{tabular}%
}
\end{table*}

\begin{table*}[]
\centering
\caption{Factuality Prompt}
\label{tab:factuality-prompt}
\resizebox{0.8\textwidth}{!}{%
\begin{tabular}{p{\linewidth}}
\hline
\textbf{»»»» Instruction »»»»} \\
You are a writing assistant who is in edit mode. You are tasked with generating edited summary based on provided a clinical note article and a model generated summary for the article. The goal is to edit the model generated summary to generate an edited summary that is factually consistent with respect to the article and contains edits introduced through an edit operation which can be one of the following: \\ \\
\textbf{Add Operation:} Intentionally add medico-legally essential words from the article to the edited summary required for accurate diagnosis and treatment documentation. Only add a single sentence in a single edit. \\ \textbf{Omit Operation:} Intentionally omit medico-legally non-essential words from the model generated summary to the edited summary not required for accurate diagnosis and treatment documentation. \\ \\
For these operations focus on words that, if present or correct in the edited summary, could lead to the right diagnoses and treatments in the future. Maintain coherence while including essential terms. The edited summary should be concise and contain no more than FIVE EXTRA WORDS compared to the model generated summary and should have an equal number of Add \& Omit operations. \\ \\
Steps for generating the edited summary: \\
\textbf{Step 1:} List the proposed edit operations to improve factually consistent in the model generated summary. \\
\textbf{Step 2:} Use the proposed edit operations to edit the model generated summary. \\
\textbf{»»»» Output Format »»»»} \\
The output format is: \\
Numbered List factuality edits made:\\
\{Edit 1\}, \{Edit 2\}, \{Edit 3\} ...\\
Edited Summary: \\
\textbf{»»»» Follow the above Instructions, Factuality Improvement Method and Output Format »»»»} \\
Now, let's start.\\ Generate the edited summary:\\
Article - \{src\}\\
Model Generated  Summary - \{ref\} \\\hline
\end{tabular}%
}
\end{table*}

\section{Human Evaluation Annotation Guidelines}
\label{appendixB}

\begin{table}[ht]
\centering

\resizebox{\columnwidth}{!}{%
\begin{tabular}{cll}
\hline
\multicolumn{3}{c}{Hallucination Instruction Identification Guideline for Add/Omit Operations} \\ \hline
\multicolumn{1}{c|}{Op.} & \multicolumn{1}{c|}{Label} & \multicolumn{1}{c}{Description} \\ \hline \hline
\multicolumn{1}{c|}{ADD} & \multicolumn{1}{l|}{0} & \begin{tabular}[c]{@{}l@{}}Including medico-legally phrases from the Article/Reference Summary \\ that are required for accurate diagnosis and treatment documentation.\end{tabular} \\ \cline{3-3}
\multicolumn{1}{c|}{ADD} & \multicolumn{1}{l|}{1} & \begin{tabular}[c]{@{}l@{}}Including medico-legally phrases from the Article/Reference Summary \\ that are not required for accurate diagnosis and treatment documentation.\end{tabular} \\ \cline{3-3}
\multicolumn{1}{c|}{OMIT} & \multicolumn{1}{l|}{0} & \begin{tabular}[c]{@{}l@{}}Not Including medico-legally phrases from the Article/Reference Summary\\ that are not required for accurate diagnosis and treatment documentation.\end{tabular} \\ \cline{3-3}
\multicolumn{1}{c|}{OMIT} & \multicolumn{1}{l|}{1} & \begin{tabular}[c]{@{}l@{}}Not Including medico-legally phrases from the Article/Reference Summary \\ that are required for accurate diagnosis andtreatment documentation.\end{tabular} \\ \hline
\end{tabular}%
}
\caption{Human annotation instructions for annotating whether an ADD/OMIT instruction is a hallucination instruction or not.}
\label{tab:human-eval-guideline-hallucination}
\end{table}

\begin{table}[ht]
\centering

\resizebox{\columnwidth}{!}{%
\begin{tabular}{cll}
\hline
\multicolumn{3}{c}{Factuality Instruction Identification Guideline for Add/Omit Operations} \\ \hline
\multicolumn{1}{c|}{Op.} & \multicolumn{1}{c|}{Label} & \multicolumn{1}{c}{Description} \\ \hline \hline
\multicolumn{1}{c|}{ADD} & \multicolumn{1}{l|}{1} & \begin{tabular}[c]{@{}l@{}}Including medico-legally phrases from the Article/Reference Summary \\ that are required for accurate diagnosis and treatment documentation.\end{tabular} \\ \cline{3-3}
\multicolumn{1}{c|}{ADD} & \multicolumn{1}{l|}{0} & \begin{tabular}[c]{@{}l@{}}Including medico-legally phrases from the Article/Reference Summary \\ that are not required for accurate diagnosis and treatment documentation.\end{tabular} \\ \cline{3-3}
\multicolumn{1}{c|}{OMIT} & \multicolumn{1}{l|}{1} & \begin{tabular}[c]{@{}l@{}}Not Including medico-legally phrases from the Article/Reference Summary\\ that are not required for accurate diagnosis and treatment documentation.\end{tabular} \\ \cline{3-3}
\multicolumn{1}{c|}{OMIT} & \multicolumn{1}{l|}{0} & \begin{tabular}[c]{@{}l@{}}Not Including medico-legally phrases from the Article/Reference Summary \\ that are required for accurate diagnosis andtreatment documentation.\end{tabular} \\ \hline
\end{tabular}%
}
\caption{Human annotation instructions for annotating whether an ADD/OMIT instruction is a factuality instruction or not.}
\label{tab:human-eval-guideline-factuality}
\end{table}

For the human evaluation, we provided the annotators with a set of clinical note articles (article), reference or unaligned model-generated summaries with the corresponding, and a list of edit instructions (edit instructions) generated by prompting \texttt{GPT-4} \& \texttt{GPT-3.5}. The edit instructions consisted of two operations (Add \& Omit operations) using which a new summary called edited summary is generated. 

The two operations in the case of \texttt{\textbf{High$\rightarrow$Low}} are described below:
\begin{enumerate}
    \item \textbf{Add Operation}: Intentionally \textbf{including} medico-legally phrases in the edited summary from the article or reference summary that \textbf{are not required} for accurate diagnosis and treatment documentation.
    \item \textbf{Omit Operation}: Intentionally \textbf{not including} medico-legally phrases in the edited summary from the article or reference summary that \textbf{are required} for accurate diagnosis and treatment documentation.
\end{enumerate}

The two operations in the case of \texttt{\textbf{Low$\rightarrow$High}} are described below:
\begin{enumerate}
    \item \textbf{Add Operation}: Intentionally \textbf{including} medico-legally phrases in the edited summary from the article or reference summary that \textbf{are required} for accurate diagnosis and treatment documentation.
    \item \textbf{Omit Operation}: Intentionally \textbf{not including} medico-legally phrases in the edited summary from the article or reference summary that \textbf{are not} required for accurate diagnosis and treatment documentation.
\end{enumerate}

Both the above operations in the edit instruction in \texttt{\textbf{High$\rightarrow$Low}} \& \texttt{\textbf{Low$\rightarrow$High}} can be used to generate hallucinations \& factuality improvements in the edited summary respectively. Here hallucinations are the phrases that are either (1) not present in the edited summary that is crucial for accurate diagnosis and treatment documentation, or (2) present in the edited summary that are not crucial for accurate\ diagnosis and treatment documentation. Similarly, factuality improvements in the edited summary are either by (1) the addition of phrases that are not present in the original summary but are crucial for accurate\ diagnosis and treatment documentation or (2) the omission of phrases that are present in the original summary by are not crucial for accurate\ diagnosis and treatment documentation. Edit instruction that leads to hallucinations is referred to as hallucination instruction, whereas edit instructions that lead to factuality improvements are referred to as factuality instructions.

The conditions for an edit instruction with either ADD or OMIT operation is a hallucination instruction is listed in Table \ref{tab:human-eval-guideline-hallucination}, and the conditions for an edit instruction with either ADD or OMIT operation is a factulaity instruction is listed in Table \ref{tab:human-eval-guideline-factuality}. In these tables, the hallucination label is used to label if an instruction leads to hallucination in the edited summary or not (0=Hallucination instruction, 1=Not a hallucination instruction), and the factuality label is used to label if an instruction leads to factuality improvements in the edited summary or not (1=Factuality instruction, 0=Not a factuality instruction).

Given the article, reference/unaligned model-generated summary, and edit instructions generated by our pipeline, we asked the annotators to annotate each instruction with its hallucination/factuality label along with a justification comment for the annotation.

For preference analysis, we also asked the annotator to give a preference label to each of the reference - edited summary pair in the case of \texttt{\textbf{High$\rightarrow$Low}}, and unaligned model generated - edited summary pair in the case of \texttt{\textbf{Low$\rightarrow$High}}. A preference label was given for each of these summary pairs as follows:

\texttt{\textbf{High$\rightarrow$Low:}} 
\begin{itemize}
    \item \texttt{Preference Label: 0} if the annotator would prefer the reference summary over the edited summary as the discharge instructions for the corresponding article.
    \item \texttt{Preference Label: 1} if the annotator would prefer the edited Summary over the reference summary as the discharge instructions for the corresponding article.
\end{itemize}

\texttt{\textbf{Low$\rightarrow$High:}}
\begin{itemize}
    \item \texttt{Preference Label: 0} if the annotator would prefer the unaligned model-generated summary over the edited Summary as the discharge instructions for the corresponding article.
    \item \texttt{Preference Label: 1} if the annotator would prefer the editted Summary over the unaligned model-generated summary as the discharge instructions for the corresponding article.
\end{itemize}

\begin{table}[]
\centering
\resizebox{\columnwidth}{!}{%
\begin{tabular}{c|cc}
\hline
 \multicolumn{3}{c}{Edit Instruction Agreement}                  \\ \hline
\hline
        & \multicolumn{2}{c}{Edit Data Setting}                  \\ \hline
Expert &
  \multicolumn{1}{c|}{\begin{tabular}[c]{@{}c@{}}High to Low\\ (Annotators agree to Hallucination)\end{tabular}} &
  \begin{tabular}[c]{@{}c@{}}Low to High\\ (Annotators agree to Factuality)\end{tabular} \\ \hline
GPT-4   & \multicolumn{1}{c|}{0.13}          & 0.30          \\
GPT-3.5 & \multicolumn{1}{c|}{\textbf{0.73}} & \textbf{0.59} \\ \hline
\hline
 \multicolumn{3}{c}{Prefrence Agreement}                  \\ \hline
         & \multicolumn{2}{c}{Edit Data Setting}                  \\ \hline
Expert &
  \multicolumn{1}{c|}{\begin{tabular}[c]{@{}c@{}}High to Low\\ (Annotators agree to Hallucination)\end{tabular}} &
  \begin{tabular}[c]{@{}c@{}}Low to High\\ (Annotators agree to Factuality)\end{tabular} \\ \hline
GPT-4   & \multicolumn{1}{c|}{0.73}          & \textbf{0.37}          \\
GPT-3.5 & \multicolumn{1}{c|}{\textbf{0.74}} & 0.33 \\ \hline
\end{tabular}%
}
\caption{Inter-annotator agreement (mean kappa score between annotations) for hallucination/factuality edits annotation.}
\label{tab:edit-inter-annotator-aggrement}
\end{table}

We also report the Kappa scores for both edit-instruction and preference label annotation used in our human evaluation, to further validate the quality of edits generated by our pipeline. Table \ref{tab:edit-inter-annotator-aggrement} reports the mean kappa score between all the annotations (1) of the edit instructions from our human evaluation samples, where each annotator gives a hallucination/factuality label ($0$ or $1$) to an edit instruction to annotate if the edit instructions leads to a hallucination/factuality edit; (2) of the preference label ($0$ or $1$) given between the edit summaries and ground truth summaries by our annotators. We observe an overall high agreement between the annotators, where edit instructions were observed to have an higher agreement when GPT-3.5 was used as the expert for both the type of edits, whereas in the case of preference labels overall \texttt{\textbf{High$\rightarrow$Low}} edit had an higher agreement for both the experts. 
\section{Edit Instruction Categories} \label{appendixc}
\begin{table}[]
\centering
\caption{Hallucination Edit Types}
\resizebox{0.8\columnwidth}{!}{%
\begin{tabular}{cll}
\hline
% \multicolumn{3}{c}{Instruction Category Labeling Guideline} \\ \hline
\multicolumn{1}{c|}{\begin{tabular}[c]{@{}c@{}}Instruction\\ Abbrivation\end{tabular}} & \multicolumn{2}{c}{Description} \\ \hline \hline
\multicolumn{1}{c|}{AR} & \multicolumn{2}{l}{Add from Reference Summary } \\
\multicolumn{1}{c|}{AA} & \multicolumn{2}{l}{Add from Article} \\
\multicolumn{1}{c|}{OR} & \multicolumn{2}{l}{Omit from Reference Summary} \\
\multicolumn{1}{c|}{OA} & \multicolumn{2}{l}{Omit from Article} \\ \hline 
\end{tabular}
}
\label{tab:edit_types}
\end{table}
From our evaluation, we observed that there were
majorly 4 types of edits made from our pipeline as shown in Table \ref{tab:edit_types}. These edits are categorized
mainly based on (1) the operation used for the edit (ADD/OMIT), and (2) whether the edit was made using the contents from the reference summary (or unaligned model generated summary) or the article (clinical note).

In order to analyze the type of edits, we promoted \texttt{GPT-4} to categorize each generated instruction into one of the edit type categories listed in Table \ref{tab:edit_types}. The prompt used is shown in Table \ref{tab:edit-type-prompt}.

For Figure \ref{fig:edit_category_analysis}, we first used the above prompt to categorize each generated instruction into one of the edit-type categories a identify the percentage-wise contribution of each edit type over all the generated instructions (top plot in Figure \ref{fig:edit_category_analysis}). Then we used the human annotation labels to calculate the percentage of instructions of each type responsible for generating the desired (hallucination/factuality improvement) outcome (bottom plot in Figure \ref{fig:edit_category_analysis}).

\begin{table*}[]
\centering
\caption{G-Eval prompt}
\label{tab:g-eval}
\resizebox{0.8\textwidth}{!}{%
\begin{tabular}{p{\linewidth}}
    \hline
    You will be given one discharge summary written for a Clinical Note. \\
    Your task is to rate the summary on one metric.
    Please make sure you read and understand these instructions carefully. Please keep this document open while reviewing, and refer to it as needed. \\
    \textbf{Evaluation Criteria:}\\
    Factual Consistency (1-10): Is the summary has missing or incorrect facts that are not supported by the source text and could lead to wrong diagnoses and treatments? \\
    \textbf{Evaluation Steps:}\\
    \begin{enumerate}
        \item Read the clinical note carefully and identify the main topic and key points.
        \item Read the discharge summary and compare it to the clinical notee. Check if the summary covers the main topic and key points of the  clinical note, and Is the summary has missing or incorrect facts that are not supported by the source text and could lead to wrong diagnoses and treatments?
        \item Assign a score for Factual Consistency on a scale of 1 to 10, where 1 is the lowest and 10 is the highest based on the Evaluation Criteria.
    \end{enumerate}
    \textbf{Clinical Note Text:}\\
    \{Document\}\\
    \textbf{Reference Discharge Summary:}\\
    \{Reference Summary \}\\
    \textbf{System Output Discharge Summary:}\\
    \{System Output Summary\}\\ \\
    Return the scores as dictionary objects, adhering to the following structure:\\
    \{"Factual Consistency": ...\} \\ 
    Please provide your response solely in the dictionary format without including any additional text. \\
    \hline
\end{tabular}%
}
\end{table*}

\begin{table*}[]
\centering
\caption{Edit type categorization prompt}
\label{tab:edit-type-prompt}
\resizebox{0.8\textwidth}{!}{%
\begin{tabular}{p{\linewidth}}
\hline
You would be given an Article, a reference summary and an edited summary along with some edit instructions.\\
    An article is a clinical note and the reference summary is a summarisation of the clinical note.\\
    Article: {{Article}}\\
    Reference Summary: {{Ref_sum}}\\
    Edited Summary: {{edit_sum}}\\
    Instructions: {{ins_truc}}\\
    The "edited summary" is generated from the "Article" and the "reference summary" using the Instructions.\\
    Instruction Types: AR, AA, OR, OA\\
    Task:\\
    Your task is to act as an expert evaluator and categorize each instruction to exactly one of the above 4 instruction types.\\
    Refer to the evaluation guide and the rules:\\
    If there are no instructions. then the counts for each category would be zero.\\
    Evaluation Guide:\\
    Step 1: Classify each instruction to either an Add instruction or an Omit Instruction.\\
    Step 2: Focus at the information mentioned in the instruction and find out, which paragraph among the reference summary and the article contains this information.\\
    Step 3: If an exact match is not found, find which paragraph among the two contains a contextually similar content.\\
    Step 4: Categorize the instructions using the rules.\\
    Step 5: Output the count of each type of instruction in the \\
    format: AR: <number>, AA: <number>, OR: <number>, OA: <number>\\
    Rules:\\
    - An instruction is to be categorized as AR if the instruction type is "Add" and if similar information can be found in the reference summary or if the instruction clearly tries to add some information before or after an instruction present in the reference summary.\\
    - An instruction is to be categorized as AA if the instruction type is "Add" and if similar information can be found in the article.\\
    - An instruction is to be categorized as OR if the instruction type is "Omit" and if similar information can be found in the reference summary.\\
    - An instruction is to be categorized as OA if the instruction type is "Omit" and if similar information can be found in the article. \\ \hline
\end{tabular}%
}
\end{table*}

\section{G-Eval Factuality Metric Prompt} \label{appendixD}
The prompt for our G-Eval evaluation is given in Table \ref{tab:g-eval}.

\section{Human Evaluation Examples}
\label{appendixE}
Examples of our human annotation are given in Table \ref{tab:example1} \& Table \ref{tab:example2}.

% \label{appendixE}
% \begin{figure}[ht]
%   \centering
%     \includegraphics[width=0.7\columnwidth]{Images/AnnotationKappa.png}
%   \caption{Kappa score for all annotation documents (ANN \#) used in the human evaluation}
%   \label{fig:annotate}
% \end{figure}

% For our human evaluation, we also calculated the Kappa score for inter-annotator agreement for hallucination labels in each document annotated by our annotators. We observed a mean Kappa score of 0.38, and the Kappa score plot for each of the documents is shown in Figure \ref{fig:annotate}. We observe that for the majority of the documents, the annotators were in high agreement with each other for the hallucination label, but in some documents in which the ADD/OMIT operation was being done using the contents of the article instead of the reference, there was a high degree of disagreement for deciding the hallucination label. We provide two examples for our human annotation in Table \ref{tab:example1} \& Table \ref{tab:example2}. 

% \begin{figure}[h]
% \centering

\section{External Evaluation Experimental setting}
\label{appendixf}
In this paper,
We trained GPT 2 and Llama 2 on the summarization dataset with 3 epochs (batch size of 8). For GPT 2, the experiments take about 2 hours. For Llama 2, the experiments take about 20 hours.
We did all the experiments with 4 NVIDIA Tesla RTX8000 GPU - 48 GB memory, with Adam optimizer -- betas=(0.9,0.999), epsilon=1e-08, learning rate=1e-04.
In all our summary generation, we used a beam size of 4, no-repeat-ngram-size=2, and minimum length and maximum length of sentences were set as (10, 256). 
We used five different random seeds to sample training data for all our experiments, and the scores reported in the tables are the average of these random seeds.

\begin{table}%[h]
\centering
\scalebox{1}{
\begin{tabular}{c|cccc}
\hline
\multirow{2}{*}{} & \footnotesize{ROUGEL} & \footnotesize{UMLS-F1}
\\

 & \multicolumn{2}{c}{GPT-3.5 Synthetic Dataset}
\\
\hline

SFT &  \footnotesize{38.03} & \footnotesize{35.47}
\\

DPO &  \footnotesize{36.48} & \footnotesize{34.51}
\\

SALT &  \footnotesize{34.87} & \footnotesize{32.84}
\\
\hline

\hline

& \multicolumn{2}{c}{GPT-4 Synthetic Dataset}
\\
\hline

SFT &  \footnotesize{38.03} & \footnotesize{35.47}
\\

DPO &  \footnotesize{34.92} & \footnotesize{32.95}
\\

SALT &  \footnotesize{34.70} & \footnotesize{32.04}
\\

\hline
\end{tabular}
}
% \vspace{-2mm}
\caption{Llama2-L2H} 

\label{table:L2H_llama2_results}
\end{table}

\clearpage
\onecolumn

\begin{longtable}{ p{14.6cm} }
\caption{Human Annotation for \texttt{\textbf{High$\rightarrow$Low}}} \\ % needs to go inside longtable environment
    % \hline
    % Human Annotation No. 1   \\ 
    \hline
    Clinical Note  \\ \hline
    \textbf{Brief Hospital Course:} Patient was found to have blood loss anemia (HCT 40s --> 22) for which she was recuscitated in ED, received 4Units PRBC, 4 FFP, 2 mg vit K, 14 mg morphine, 1 mg ativan, 1600 cc NS. A foley was placed which revealed frank blood. She also had ARF (BUN/Cr 21/1.8). Abdomenal U/S, CT and cystoscopy revealed prevesicular and intrabladder hematomas in the context of anticoagulation along with blood loss into right lower extremity. The hematomas were evacuated (urology and vascular [**Doctor First Name **]), and an intra-op right ureteral stent was placed for her obstructive uropathy w/ right hydronephrosis. Pt developed post-surgical ileus that resolved in two days. RLE U/S was negative for IVC thrombus or LE venous compression. Intial ARF was likely pre-renal secondary to major volume loss and resolved with fluid resusitation. The ureter stent was removed once her Cr normalized. In transitioning her back to coumadin via heparin, she developed HIT (Plt 200s -----> 23 over 2 days). Given her high risk for thrombosis (PAF, Obese, immobile) she was initially treated with Argatroban (Per heme recs, Dr. [**Last Name (STitle) **]. However, she became jaundice w/ elevated bilirubin. It was reasoned that this could possibly be secondary to Argatroban liver toxicity in addition to low grade hemolysis (resorbing residual hematoma). A RUQ U/S failed to show any changes in her liver, biliary tree, or gallbladder. She was ultimately switched to Lepirudin for anticoagulation, bridged to coumadin when platellet levels reached 100,000, and obtained a theurapeutic INR before discharge.
    \\ \hline
    Reference Summary   \\ \hline
    \textbf{Discharge Instructions:} \\
    1. If you develop chest pain, shortness of breath, palpitations, lightheadedness, abdominal pain develop seek immediate medical attention.\\ 
    2. Take all medications as listed below. A number of medications have been changed so please review the list carefully. Check with Dr. [**Last Name (STitle) 5310**] concerning when to restart your aspirin. \\
    3. Follow up as directed below. \\
    4. You will need an INR checked in 2 days with instructions on adjusting your coumadin dose by Dr. [**Last Name (STitle) 5310**] \\
    5. You have been provided with a note excusing you from work for the next month. Please contact your PCP for instructions on when to return after that."
\\ \hline
    GPT-4 Generated Edit Instructions   \\ \hline
    Numbered List hallucination edits made: \\
    \begin{enumerate}
        \item Add Operation: Add "blood loss anemia" to the summary.
        \item Omit Operation: Omit "shortness of breath" from the summary.
        \item Add Operation: Add "HIT" to the summary.
        \item Omit Operation: Omit "palpitations" from the summary.
        \item Add Operation: Add "Argatroban liver toxicity" to the summary.
    \end{enumerate}
    \\ \hline
    GPT-4 Generated Edited Summary   \\ \hline
    \textbf{Discharge Instructions:} 
    
    1. If you develop chest pain, lightheadedness, abdominal pain or symptoms of blood loss anemia and HIT, seek immediate medical attention. \\
    2. Take all medications as listed below. A number of medications have been changed so please review the list carefully. Check with Dr. [**Last Name (STitle) 5310**] concerning when to restart your aspirin and in case of symptoms of Argatroban liver toxicity. \\
    3. Follow up as directed below. \\
    4. You will need an INR checked in 2 days with instructions on adjusting your coumadin dose by Dr. [**Last Name (STitle) 5310**] \\
    5. You have been provided with a note excusing you from work for the next month. Please contact your PCP for instructions on when to return after that."
    \\ \hline
    Human Annotation for GPT-4 Edited Summary (by a doctor)\\ \hline
    Instruction 1: \\
    Hallucination Label: 0 \\
    Comment: a description \\
    Instruction 2: \\
    Hallucination Label: 1 \\
    Comment: patient may ignore this  \\
    Instruction 3: \\
    Hallucination Label: 0 \\
    Comment: nothing changes  \\
    Instruction 4: \\
    Hallucination Label: 1 \\
    Comment: patient may ignore this  \\
    Instruction 5:\\
    Hallucination Label: 0 \\
    Comment: nothing changes \\
    \\ \hline
    GPT-3.5 Generated Edit Instructions \\ \hline
    Numbered List hallucination edits made: \\
    \begin{enumerate}
        \item Add Operation: "Please note that" before "If you develop chest pain" 
        \item Omit Operation: Omit "shortness of breath"
        \item Add Operation: "and" before "palpitations"
        \item Omit Operation: Omit "lightheadedness"
        \item Add Operation: "and" before "abdominal pain" 
        \item Add Operation: "Please" before "review the list carefully" 
        \item Add Operation: "Please" before "Check with Dr. [**Last Name (STitle) 5310**]"
        \item Add Operation: "Please" before "Follow up as directed below"
        \item Add Operation: "Please" before "You will need an INR checked in 2 days"
    \end{enumerate} \\
    \\ \hline
    GPT-3.5 Generated Edited Summary \\ \hline
    \textbf{Discharge Instructions:} Please note that if you develop chest pain and palpitations and abdominal pain develop seek immediate medical attention. Take all medications as listed below. Please review the list carefully. Please Check with Dr. [**Last Name (STitle) 5310**] concerning when to restart your aspirin. Please Follow up as directed below. Please You will need an INR checked in 2 days with instructions on adjusting your coumadin.
    \\ \hline
    Human Annotation for GPT-3.5 Edited Summary (by a doctor)\\ \hline
    Instruction 1: \\
    Hallucination Label: 0 \\
    Comment:  \\
    Instruction 2: \\
    Hallucination Label: 1 \\
    Comment: this is important  \\
    Instruction 3: \\
    Hallucination Label: 0 \\
    Comment: \\
    Instruction 4: \\
    Hallucination Label: 1 \\
    Comment: import disease  \\
    Instruction 5:\\
    Hallucination Label: 0 \\
    Comment:  \\
    Instruction 6:\\
    Hallucination Label: 0 \\
    Comment:  \\
    Instruction 7:\\
    Hallucination Label: 0 \\
    Comment:  \\
    Instruction 8:\\
    Hallucination Label: 0 \\
    Comment:  \\
    Instruction 9:\\
    Hallucination Label: 0 \\
    Comment:  \\ \hline
\label{tab:example1}
\end{longtable}

\begin{longtable}{ p{14.6cm} }
\caption{Human Annotation for \texttt{\textbf{Low$\rightarrow$High}}} \\ % needs to go inside longtable environment
    % \hline
    % Human Annotation No. 1   \\ 
    \hline
    Clinical Note  \\ \hline
    \textbf{Brief Hospital Course:} As mentioned in the HPI, pt was initially seen by cardiac surgery following his cardiac cath. Patient was eventually consented for surgery and on [**2155-2-13**] he was brought to the operating room where he underwent a coronary artery bypass graft x 3 and aortic valve replacement. Please see op note for surgical details. Patient tolerated the procedure well and was transferred to the CSRU in stable condition receiving Neo-Synephrine, Dobutamine, and Levophed drips. Later on op day sedation was weaned and patient awoke neurologically intact. He was then weaned from mechanical ventilation and extubated. He was weaned from all Inotropes/Pressors on post op day one and was then transferred to the cardiac step-down unit. B Blockers and Diuretics were initiated and patient was gently diuresed towards his pre-op weight. Chest tubes and Foley catheter were removed on post op day two. And epicardial pacing wires were removed on post op day three. Pt was followed by physical therapy during his entire post op course for strength and mobility. Patient had a relatively uncomplicated post op course and was at level 5 by post op day five. His exam and labs were stable on post op day five and he was discharged home with VNA services and the appropriate follow-up appointments.
    \\ \hline
    Unaligned Model Generated Summary   \\ \hline
    \textbf{Discharge Instructions:} [**Month (only) 116 **] shower, no baths or swimming Monitor wounds for infection redness, drainage, or increased pain Report any fever greater than 101 Report any weight gain of greater than 2 pounds in 24 hours or 5 pounds in a week No creams, lotions, powders, or ointments to incisions No driving for approximately one month No lifting more than 10 pounds for 10 weeks Please call with any questions or concerns.
\\ \hline
    GPT-4 Generated Edit Instructions   \\ \hline
    Numbered List hallucination edits made: \\
    \begin{enumerate}
        \item Add Operation: Add "Patient underwent a coronary artery bypass graft x 3 and aortic valve replacement." from the article to the model generated summary.
        \item Omit Operation: Omit "No creams, lotions, powders, or ointments to incisions" from the model generated summary.
    \end{enumerate}
    \\ \hline
    GPT-4 Generated Edited Summary   \\ \hline
    \textbf{Discharge Instructions:} 
    
    Patient underwent a coronary artery bypass graft x 3 and aortic valve replacement. Shower, no baths or swimming. Monitor wounds for infection - redness, drainage, or increased pain. Report any fever greater than 101. Report any weight gain of greater than 2 pounds in 24 hours or 5 pounds in a week. No driving for approximately one month. No lifting more than 10 pounds for 10 weeks. Please call with any questions or concerns.
    \\ \hline
    Human Annotation for GPT-4 Edited Summary (by a doctor)\\ \hline
    Instruction 1: \\
    Factuality Label: 1 \\
    Comment: useful \\
    Instruction 2: \\
    Hallucination Label: 0 \\
    Comment: This is not useful  \\
    \\ \hline
    GPT-3.5 Generated Edit Instructions \\ \hline
    Numbered List hallucination edits made: \\
    \begin{enumerate}
        \item Add Operation: "Patient underwent a coronary artery bypass graft x 3 and aortic valve replacement." 
        \item Omit Operation: "No creams, lotions, powders, or ointments to incisions." 
        \item Add Operation: "Patient was transferred to the cardiac step-down unit." 
        \item Omit Operation: "No driving for approximately one month." 
        \item Add Operation: "Patient was discharged home with VNA services and the appropriate follow-up appointments." 
    \end{enumerate} \\
    \\ \hline
    GPT-3.5 Generated Edited Summary \\ \hline
    \textbf{Discharge Instructions:} [**Month (only) 116**] shower, no baths or swimming. Monitor wounds for infection - redness, drainage, or increased pain. Report any fever greater than 101. Report any weight gain of greater than 2 pounds in 24 hours or 5 pounds in a week. No lifting more than 10 pounds for 10 weeks. Please call with any questions or concerns. Patient underwent a coronary artery bypass graft x 3 and aortic valve replacement. Patient was transferred to the cardiac step-down unit. Patient was discharged home with VNA services and the appropriate follow-up appointments.
    \\ \hline
    Human Annotation for GPT-3.5 Edited Summary (by a doctor)\\ \hline
    Instruction 1: \\
    Hallucination Label: 1 \\
    Comment: Knowing what surgeries were performed is important to patient.\\
    Instruction 2: \\
    Hallucination Label: 0 \\
    Comment: This is common sense like notification, but I think this is also importatnt.  \\
    Instruction 3: \\
    Hallucination Label: 0 \\
    Comment: This explanation sounds not helpful in discharge note.\\
    Instruction 4: \\
    Hallucination Label: 0 \\
    Comment: Be careful with all kinds of risks.  \\
    Instruction 5:\\
    Hallucination Label: 1 \\
    Comment: A notification for future plan\\ \hline
\label{tab:example2}
\end{longtable}

\clearpage
\twocolumn

\end{document}